\documentclass[letterpaper]{article} 
\usepackage{aaai24}  
\usepackage{times}  
\usepackage{helvet}  
\usepackage{courier}  
\usepackage[hyphens]{url}  
\usepackage{graphicx} 
\usepackage{natbib}  
\usepackage{caption} 
\usepackage{algorithm}
\usepackage{algorithmicx}

\usepackage{newfloat}
\usepackage{listings}
\DeclareCaptionStyle{ruled}{labelfont=normalfont,labelsep=colon,strut=off} 
\floatstyle{ruled}
\newfloat{listing}{tb}{lst}{}
\floatname{listing}{Listing}
\usepackage{amsmath}
\usepackage{amssymb}
\usepackage{mathtools}
\usepackage{caption}
\usepackage{subcaption}
\usepackage{relsize}
\usepackage{algpseudocode}
\usepackage{color}
\usepackage{multirow}
\newtheorem{theorem}{Theorem}
\newtheorem{lemma}[theorem]{Lemma}

\title{On Inference Stability for Diffusion Models}
\author{
    Viet Nguyen\textsuperscript{\rm 1}\equalcontrib\thanks{This work was partly done while at HUST}, Giang Vu\textsuperscript{\rm 2,3}\equalcontrib, Tung Nguyen Thanh\textsuperscript{\rm 2,3}, Khoat Than\textsuperscript{\rm 2}\thanks{Corresponding author}, Toan Tran\textsuperscript{\rm 1}\\
} 
\affiliations{
    \textsuperscript{\rm 1}VinAI Research, Vietnam\\
    \textsuperscript{\rm 2}Hanoi University of Science and Technology, Vietnam\\
    \textsuperscript{\rm 3} Viettel Group, Vietnam\\
    \{v.vietnv18, v.toantm3\}@vinai.io,  khoattq@soict.hust.edu.vn, \{giangvl2, tungnt759\}@viettel.com.vn
}

\begin{document}

\maketitle

\begin{abstract}
Denoising Probabilistic Models (DPMs) represent an emerging domain of generative models that excel in generating diverse and high-quality images. However, most current training methods for DPMs often neglect the correlation between timesteps, limiting the model's performance in generating images effectively. Notably, we theoretically point out that this issue can be caused by the cumulative estimation gap between the predicted and the actual trajectory. To minimize that gap, we propose a novel \textit{sequence-aware} loss that aims to reduce the estimation gap to enhance the sampling quality. Furthermore, we theoretically show that our proposed loss function is a tighter upper bound of the estimation loss in comparison with the conventional loss in DPMs. Experimental results on several benchmark datasets including CIFAR10, CelebA, and CelebA-HQ consistently show a remarkable improvement of our proposed method regarding the image generalization quality measured by FID and Inception Score compared to several DPM baselines. Our code and  pre-trained checkpoints are available at \url{https://github.com/VinAIResearch/SA-DPM}.
\end{abstract}

\section{Introduction}

Diffusion Probabilistic Models (DPMs) \cite{sohl2015deep}, inspired by statistical physics, have been shown to be more effective generative models than prior ones. Typically, a DPM consists of two processes: a forward process that gradually adds noise to the original data distribution and a reverse process that learns to iteratively reconstruct a data instance from the noises. As a progress of that idea, \cite{ho2020denoising} proposes Denoising Diffusion Probabilistic Models (DDPMs) which exploit the knowledge about the transition distribution to derive the loss function and guide the training process. Parallel to that work, \cite{song2019generative} uses the score-based model to train a similar model. More recently, \cite{song2021scorebased} interprets those two works under the lens of stochastic differential equations. This class of models outperforms prior ones in terms of generated images' quality and distribution coverage. While other likelihood-based generative models require unique assumptions on data \cite{pmlr-v37-germain15, van2016conditional} or constraints in model architecture \cite{dinh2017density, papamakarios2017masked, kingma2018glow, ho2019flow} to perform well, DPMs do not hold any of that requirements. Moreover, compared with Generative Adversarial Networks, Diffusion Models do not require adversarial training thus making the learning process easy and stable.

Although DPMs have been shown to achieve state-of-the-art results in various data generation tasks since their debut, these models often suffer from slow sampling speed, which may require thousands of model feeds to achieve high sample quality. To address this issue, many researchers have focused on accelerating the generating process. For example, \cite{song2020denoising, kong2021on} propose non-Markovian diffusion processes, which allow taking multiple steps at once to accelerate the sampling time. Several works explore finding short sampling trajectories by applying search algorithms, e.g., grid search \cite{chen2021wavegrad}, dynamic programming 
\cite{watson2021learning}, and differentiable search \cite{watson2022learning}. \cite{salimans2022progressive, song2023consistency} propose to boost the sampling process via knowledge distillation with the core idea of distilling a multi-step process into a single step. 

\cite{song2021scorebased} establishes a connection between the denoising process and solving ordinary differential equations (ODE). Such a connection enables the use of numerical methods of differential equations to accelerate the denoising process. While \cite{song2021scorebased} proposes the use of higher-order solvers such as Runge-Kutta methods, \cite{liu2022pseudo} proposes
pseudo-numerical methods to generate samples along a specific manifold. Another approach proposed by \cite{Karras2022edm} is to use Heun’s second-order method to solve the probability flow ODE. 

Some recent attempts aim to refine inefficient sampling trajectories due to the approximation and optimization errors in training. \cite{bao2022analyticdpm, bao2022estimating} propose to estimate the optimal variance to correct the potential bias caused by the imperfect mean estimation. Meanwhile, \cite{Zhang_2023_CVPR} introduces an extrapolation operation on two consecutive sampling steps to make the sampling trajectory closer to the direction of the real-data point. 

One main drawback of those works is that they mostly focus on sampling efficiency by, for instance, making modifications in only the sampling process, or fine-tuning pre-trained DPMs, without training DPMs from scratch. In particular, we find out that most existing DPMs are often trained in a timestep-independence paradigm, which often ignores the sequential nature of DPMs in both forward and backward processes. We view the sampling trajectory at a global scale and derive the estimation gap of a noise predictor. That gap indicates how far the predicted trajectory is from the actual one. From that observation, we propose a new training objective, termed the \textit{Sequence-Aware} (\textbf{SA}) loss, that constrains directly  the gap. Our contributions are summarized below:

\begin{itemize}
    \item We point out the estimation gap between the predicted and actual sampling trajectory and analyze its effect on the data generation quality of DPMs.  
    \item We propose a novel \textit{sequence-aware loss} and an induced training algorithm to minimize the estimation gap.
    \item We theoretically show that our loss function is a tighter upper bound of the estimation gap in comparison with the conventional loss function.
    \item We employ that loss in multiple DPM baselines. Empirical results illustrate significant improvements in FID and Inception Score compared to several current DPM baselines.
\end{itemize}

\section{Background}

Diffusion Probabilistic Models \cite{sohl2015deep} are comprised of two fundamental components, including the \textit{forward process} and the \textit{reverse process}. The former gradually diffuses each input $\boldsymbol{x}_0$, following a data distribution $q(\boldsymbol{x}_{0})$, into a standard Gaussian noise through $T$ timesteps, i.e., $\boldsymbol{x}_T \sim \mathcal N(\mathbf{0}, \mathbf{I})$, where $\mathbf{I}$ is the identity matrix, $\mathcal{N}(\cdot,\cdot)$ represents the normal distribution.  The reverse process starts from $\boldsymbol{x}_T$ and then interactively denoises to get an original image. We recap the background of DPMs following the idea of DDPM \cite{ho2020denoising}.

\subsection{Forward Process}

Given an original data distribution $q(\boldsymbol{x}_{0})$, the forward process can be presented as follows:
\begin{align*}
    &q(\boldsymbol{x}_{1:T}|\boldsymbol{x}_{0}) = \prod_{t=1}^{T} q(\boldsymbol{x}_{t}|\boldsymbol{x}_{t-1}),
\end{align*}
where $q(\boldsymbol{x}_{t}|\boldsymbol{x}_{t-1}) \coloneqq \mathcal{N}(\boldsymbol{x}_{t}; \sqrt{1-\beta_{t}}\boldsymbol{x}_{t-1}, \beta_{t}\mathbf{I})$ and an increasing noise scheduling sequence $\beta_{t} \in (0, 1]$, which describes the amount of noise added at each timestep $t$. Denoting $\alpha_{t} = 1 - \beta_{t}$ and $\bar{\alpha}_{t} = \prod_{s=1}^{t}\alpha_{s}$, the distribution of diffused image $\boldsymbol{x}_t$ at timestep $t$ has a closed form as: 
\begin{align*}
q(\boldsymbol{x}_{t}|\boldsymbol{x}_{0}) = \mathcal{N}(\boldsymbol{x}_{t}; \sqrt{\bar{\alpha}_{t}}\boldsymbol{x}_{0}, (1-\bar{\alpha}_{t})\mathbf{I}).
\end{align*}

By applying the reparameterization trick~\cite{kingma2013auto, rezende2014stochastic}, we can sample the data at each time step $t$ by:
\begin{equation}\label{eq:xt_from_x0}
    \boldsymbol{x}_{t} = \sqrt{\bar{\alpha}_{t}}\boldsymbol{x}_{0} + \sqrt{1-\bar{\alpha}_{t}}\boldsymbol{\epsilon}_t,
\end{equation}
where $\boldsymbol{\epsilon}_{t} \sim \mathcal{N}(\mathbf{0}, \mathbf{I})$. The noise scheduler $\beta_{1:T}$ is designed in such a way that $\bar{\alpha}_{1:T}$ is a decreasing array and $\bar{\alpha}_{T} \approx 0$. That means at the end of the forward process, $\boldsymbol{x}_{T}$ is likely sampled from the standard Gaussian distribution $\mathcal{N}(\mathbf{0}, \mathbf{I})$.

\begin{algorithm}[t]
\caption{Conventional training}\label{alg:convential_training}
\begin{algorithmic}
\Require Empirical data distribution $q$, number $T$ of timesteps, the noise predictor $\boldsymbol{f}_{\theta}$, learning rate $\eta$. 
\Repeat
\State $\boldsymbol{x}_{0} \sim q(\boldsymbol{x}_{0})$
\State $t \sim \mathrm{Uniform}(\{1, \dots, T\})$
\State $\epsilon \sim \mathcal{N}(\mathbf{0}, \mathbf{I})$
\State $\boldsymbol{x}_{t} = \sqrt{\bar{\alpha}_{t}}\boldsymbol{x}_{0} + \sqrt{1-\bar{\alpha}_{t}}\epsilon$
\State $\mathcal{L}_{simple} = \|\boldsymbol{f}_{\theta}(\boldsymbol{x}_{t}, t)-\boldsymbol{\epsilon}_{t}\|^{2}$ 
\State $\theta \leftarrow \theta - \eta \bigtriangledown_{\theta}\mathcal{L}_{simple}$
\Until converged
\end{algorithmic}
\end{algorithm}

\begin{algorithm}[t]
\caption{Sampling}
\label{alg:conventional_sampling}
\begin{algorithmic}
\State $\boldsymbol{x}_{T} \sim \mathcal{N}(\mathbf{0}, \mathbf{I})$
\State $\bar{\boldsymbol{x}}_{T} = \boldsymbol{x}_{T}$
\For{$t = T, \dots, 1$}
\State $\boldsymbol{z} \sim \mathcal{N}(\mathbf{0}, \mathbf{I})$ if $t > 1$, else $\boldsymbol{z} = 0$
\State $\bar{\boldsymbol{x}}_{t-1} = \frac{1}{\sqrt{\alpha_{t}}}\mathlarger{(} \bar{\boldsymbol{x}}_{t} - \frac{1-\alpha_{t}}{\sqrt{1-\bar{\alpha}_{t}}} \boldsymbol{f}_{\theta}(\bar{\boldsymbol{x}}_{t}, t)\mathlarger{)} + \sigma_{t}\boldsymbol{z}$
\EndFor
\State \textbf{return} $\bar{\boldsymbol{x}}_0$ 
\end{algorithmic}
\end{algorithm}

\subsection{Reverse Process}

At each step of the forward diffusion process, only a small amount of Gaussian noise is added to the data. Therefore, the reverse conditional distribution $q(\boldsymbol{x}_{t-1}|\boldsymbol{x}_{t})$ can be approximated by a Gaussian conditional distribution 
$$q(\boldsymbol{x}_{t-1}|\boldsymbol{x}_{t},\boldsymbol{x}_0)=\mathcal{N}(\boldsymbol{x}_{t-1}; \tilde{\boldsymbol{\mu}}_{t}(\boldsymbol{x}_{t}, \boldsymbol{x}_{0}), \tilde{\beta}_{t}\mathbf{I}),$$
where $\tilde{\beta}_{t} = \frac{1-\bar{\alpha}_{t-1}}{1-\bar{\alpha}_{t}} \beta_{t}$ and 
\begin{align}
    \tilde{\boldsymbol{\mu}}_{t}(\boldsymbol{x}_{t}, \boldsymbol{x}_{0}) &= \gamma_{1, t}\boldsymbol{x}_{0} + \gamma_{2, t}\boldsymbol{x}_{t}, \label{eq:posterior_mean} \\
     \gamma_{1, {t}} &= \frac{\sqrt{\bar{\alpha}_{t-1}} \beta_{t}}{1-\bar{\alpha}_{t}}, \quad 
    \gamma_{2, {t}} &= \frac{\sqrt{\alpha_{t}}(1-\bar{\alpha}_{t-1})}{1-\bar{\alpha}_{t}}. \nonumber
\end{align} 
Therefore, the trained denoising process $p_{\theta}(\boldsymbol{x}_{t-1}|\boldsymbol{x}_{t})$ to approximate $q(\boldsymbol{x}_{t-1}|\boldsymbol{x}_{t},\boldsymbol{x}_0)$ can be parameterized by
\begin{align*}
    p_{\theta}(\boldsymbol{x}_{t-1}|\boldsymbol{x}_{t}) &= \mathcal{N}(\boldsymbol{x}_{t-1}; \boldsymbol{\mu}_{\theta}(\boldsymbol{x}_{t},t), \sigma^{2}_{t}\mathbf{I}),
\end{align*}
where $\boldsymbol{\mu}_{\theta}(\boldsymbol{x}_{t},t)$ and $\sigma^{2}_{t}\mathbf{I}$ are the mean and covariance matrix of the parametric denoising model, respectively.

The training objective is then to maximize a variational lower bound on the log-likelihood of the original $\boldsymbol{x}_0$, which can be simplified (by excluding an additional term that is irrelevant to the training) as minimizing the loss:
\begin{multline*}
\mathcal{L}(\theta)=-\log p_{\theta}(\boldsymbol{x}_0|\boldsymbol{x}_1) \\
+\textstyle\sum_{t}D_{KL}(q(\boldsymbol{x}_{t-1}|\boldsymbol{x}_{t},\boldsymbol{x}_0)||p_{\theta}(\boldsymbol{x}_{t-1}|\boldsymbol{x}_{t})).
\end{multline*}

\begin{figure*}[t]
    \centering
    \begin{subfigure}{0.35\textwidth}
        \includegraphics[width=\linewidth]{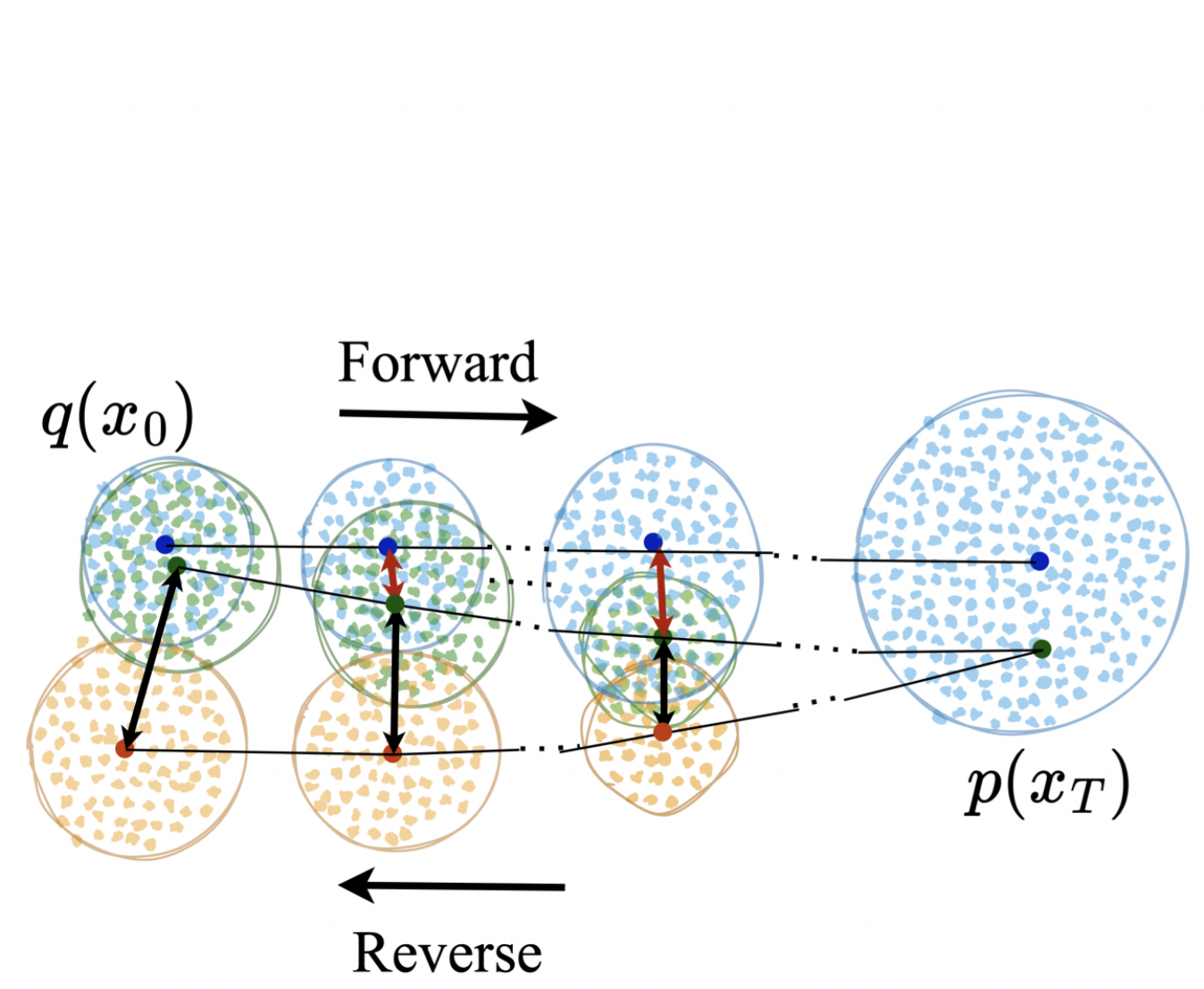}
        \caption{Large estimation gap $\bar{d}_{\theta}$}
        \label{fig:sampling_trajectory}
    \end{subfigure}
    \hfill
    \begin{subfigure}{0.15\textwidth}
        \includegraphics[width=\linewidth]{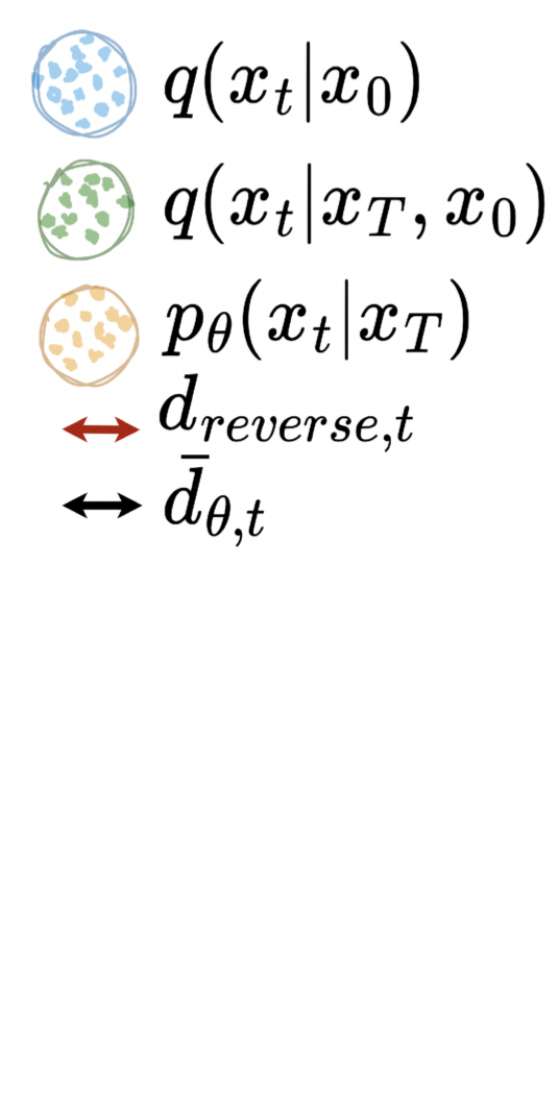}
    \end{subfigure}
    \hfill
    \begin{subfigure}{0.35\textwidth}
        \includegraphics[width=\linewidth]{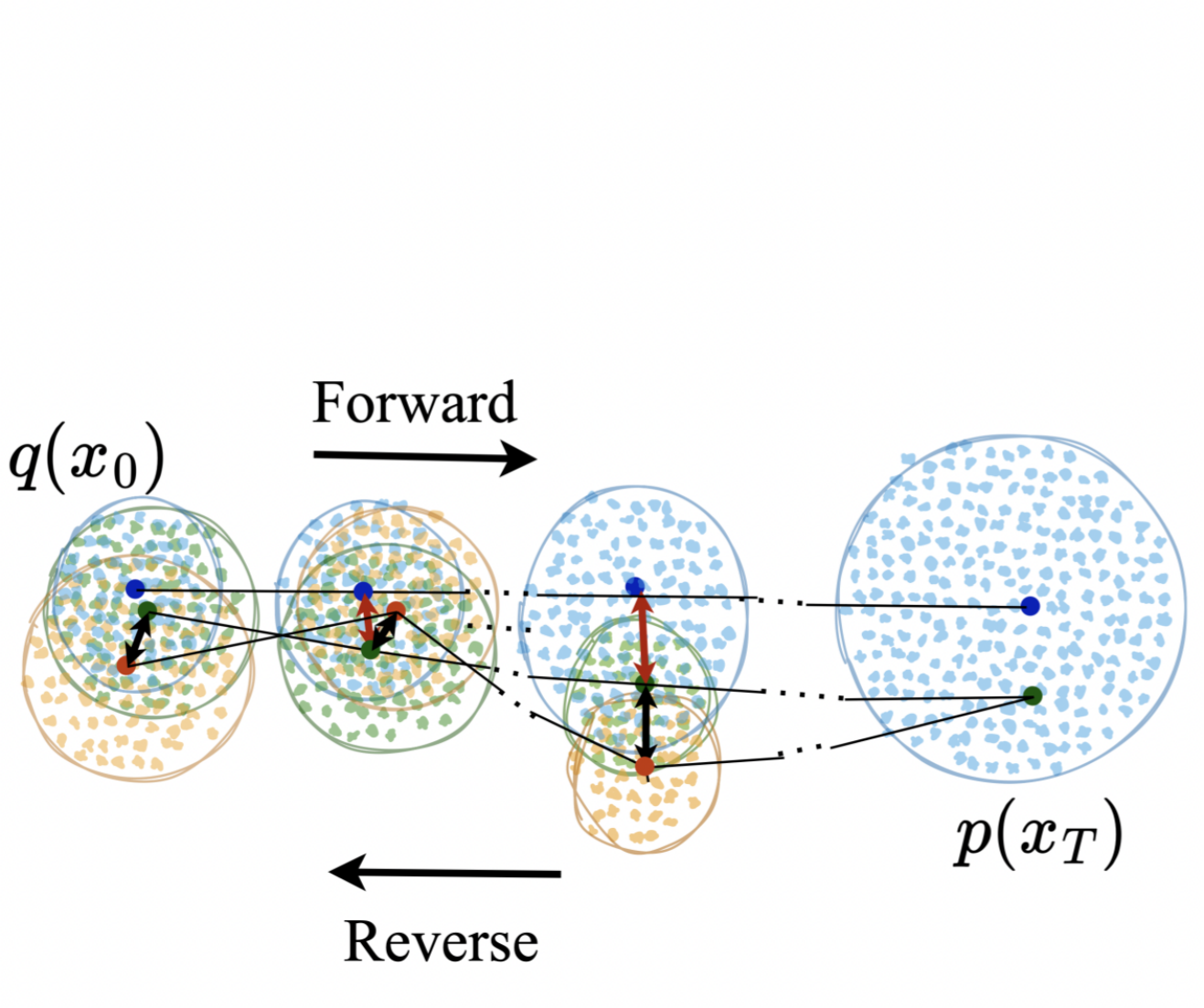}
        \caption{Small estimation gap $\bar{d}_{\theta}$}
        \label{fig:expected_sampling_trajectory}
    \end{subfigure}
    \caption{1-D example of sampling trajectory. Under the assumption that the error at each timestep is similar: (a) the cumulative error by steps is large while (b) the cumulative error by steps is small. This behavior is due to the correlation between neighbor timesteps.
    }
\end{figure*}

The mean $\boldsymbol{\mu}_{\theta}(\boldsymbol{x}_{t},t)$ predicted by the denoising model at each step can be reparameterized as a neural network  that predicts the true $\boldsymbol{x}_0$. Alternately, following \cite{ho2020denoising}, one can use a noise prediction model $\boldsymbol{f}_{\theta}$ that predicts the noise $\boldsymbol{\epsilon}_t$ added to $\boldsymbol{x}_0$ to construct $\boldsymbol{x}_t$. This allows training by simply minimizing the mean squared error between the predicted noise $\boldsymbol{f}_{\theta}(\boldsymbol{x}_t,t)$ and the true added Gaussian noise $\boldsymbol{\epsilon}_t$ (detailed in Algorithm \ref{alg:convential_training}):
\begin{align}
    \mathcal{L}_{simple} &= \mathbf{E}_{t, \boldsymbol{x}_{0}, \epsilon_{t}}\mathlarger{[}\|\boldsymbol{f}_{\theta}(\boldsymbol{x}_{t},t) - \boldsymbol{\epsilon}_{t}\|^{2}\mathlarger{]} \label{eq:conventional_loss}.
\end{align} 

After training, new  samples can be generated by first sampling Gaussian noise $\boldsymbol{x}_{T} \sim \mathcal{N}(\mathbf{0}, \mathbf{I})$ and then passing this noise through the trained model's iterative denoising procedure over $T$ timesteps, ultimately outputting a new  sample $\boldsymbol{x}_0$, detailed in Algorithm~\ref{alg:conventional_sampling}.

\section{Methodology}

In the sampling phase, a small amount of error may be introduced in each denoising iteration due to the imperfect learning process. Note that the inference process often requires many iterations to produce high-quality images, leading to the accumulation of these errors. In this section, we first point out the estimation gap between the predicted and ground-truth noises in the sampling process of DPMs and show its importance in the training phase to mitigate this accumulation and improve the quality of generated images. Based on that gap, we introduce a novel loss function that is proven to be tighter than $\mathcal L_{simple}$ commonly used  in DPMs.

\subsection{Estimation Gap}

The data generation process in Diffusion Models is performed by iteratively sampling a datapoint from the predicted distribution of $q(\boldsymbol{x}_{t-1}|\boldsymbol{x}_{t}, \boldsymbol{x}_{0})$. To interpret the working principle of the global trajectory, we take a further derivation on $q({\boldsymbol{x}_{t-1}|\boldsymbol{x}_{T}, \boldsymbol{x}_{0}})$,  detailed in Appendix A, to obtain
\begin{align*}
    q(\boldsymbol{x}_{t-1}|\boldsymbol{x}_{T}, \boldsymbol{x}_{0}) &= \mathcal{N}(\boldsymbol{x}_{t-1}; \boldsymbol{\mu}^{'}_{t}, \beta^{'}_{t}\mathbf{I}), \\
    \text{where } \boldsymbol{\mu}^{'}_{t} &= \sqrt{\bar{\alpha}_{t-1}}\boldsymbol{x}_{0} + \frac{\sqrt{\bar{\alpha}_{T}}(1-\bar{\alpha}_{t-1})}{\sqrt{\bar{\alpha}_{t-1}(1-\bar{\alpha}_{T})}} \boldsymbol{\epsilon}_{T}.
\end{align*}
Here, we can ignore the variance term since it is fixed in basic settings. We define $d_{reverse, t}=\frac{\sqrt{\bar{\alpha}_{T}}(1-\bar{\alpha}_{t-1})}{\sqrt{\bar{\alpha}_{t-1}(1-\bar{\alpha}_{T})}} \boldsymbol{\epsilon}_{T}$ as the reverse gap term. As   $\frac{\sqrt{\bar{\alpha}_{T}}(1-\bar{\alpha}_{t-1})}{\sqrt{\bar{\alpha}_{t-1}(1-\bar{\alpha}_{T})}}$ decreases to 0 when $t$ comes to the first step, the mean $\boldsymbol{\mu}^{'}_{t}$ converges to $\boldsymbol{x}_{0}$ naturally. In many real-life applications, at each timestep $t$, the sampling phase of DPMs aims to provide an approximation $\boldsymbol{x}_{\theta, 0}^{(t)}$ of the true value $\boldsymbol{x}_{0}$ and the corresponding vector error $(\boldsymbol{x}_{\theta, 0}^{(t)} - \boldsymbol{x}_{0})$ is then expected to be sufficiently close to $\mathbf{0}$. 

Technically, according to \eqref{eq:posterior_mean}, the mean of the posterior distribution $q(\boldsymbol{x}_{t-1}|\boldsymbol{x}_{t}, \boldsymbol{x}_{0})$ at each timestep $t$ is defined as: $\tilde{\boldsymbol{\mu}}_{t} = \gamma_{1, t} \boldsymbol{x}_{0} + \gamma_{2, t}\boldsymbol{x}_{t}$. Note that $\boldsymbol{x}_{t}$ does not depend on the prediction $\boldsymbol{x}_{\theta, 0}^{(t)}$. Given the true noise $\boldsymbol{\epsilon}_{1:T}$ added to $\boldsymbol{x}_{0}$, according to (\ref{eq:xt_from_x0}), the gap incurred by the noise predictor $\boldsymbol{f}_{\theta}(\boldsymbol{x}_{t}, t)$ at step $t$ is defined as: 
\begin{align}\label{eq:gap}
d_{\theta, t} = \gamma_{1, t}(\boldsymbol{x}_{\theta, 0}^{(t)} - \boldsymbol{x}_{0}) 
            =\gamma_{1, t}\frac{\sqrt{1-\bar{\alpha}_{t}}}{\sqrt{\bar{\alpha}_{t}}}(\boldsymbol{f}_{\theta}(\boldsymbol{x}_{t}, t) - \boldsymbol{\epsilon}_{t}).
\end{align}

Now we can formally point out the gap between the true noises and predictions by a model.

\begin{theorem}[Estimation gap] \label{thm-Estimation-gap}
Let $\boldsymbol{f}_{\theta}(\boldsymbol{x}_{s}, s)$ be a noise predictor with parameter $\theta$. Its total gap from step 2 to $T$, for each $\boldsymbol{x}_{0}$, is 
\begin{equation}
\label{eq:global_gap}
d_{\theta}(\boldsymbol{x}_{0}) = \sum_{i=2}^{T}\tau_{i}(\boldsymbol{f}_{\theta}(\boldsymbol{x}_{i}, i) - \boldsymbol{\epsilon}_{i}),
\end{equation}
where $\tau_{i} = \frac{\sqrt{\bar{\alpha}_{i-1}}(1-\bar{\alpha}_{1})}{\sqrt{\alpha_{1}}(1-\bar{\alpha}_{i-1})}\gamma_{1, i} \frac{\sqrt{1-\bar{\alpha}_{i}}}{\sqrt{\bar{\alpha}_{i}}}$. Furthermore, the total loss of $\boldsymbol{f}_{\theta}$ is $\mathcal{L}_{\theta} = \mathbf{E}_{\boldsymbol{x}_{0}, \boldsymbol{\epsilon}} \| d_{\theta}(\boldsymbol{x}_{0}) \|^2$.
\end{theorem}

\textit{Proof sketch. }
Denote $\bar{d}_{\theta, T} = {d}_{\theta, T}$ and define $\bar{d}_{\theta, t} = d_{\theta, t} + \gamma_{2, t} \bar{d}_{\theta, t+1}$ to be the gap at an arbitrary timestep $t < T$. By  induction (Appendix B), we have 
\begin{align}\label{eq:total_gap}
    \bar{d}_{\theta, t}
    &= d_{\theta, t} + \sum_{i=t+1}^{T} \left[\prod_{s=t}^{i-1}\gamma_{2, s}\right]d_{\theta, i} \nonumber \\
    &= d_{\theta, t} + \sum_{i=t+1}^{T} \left[ \frac{\sqrt{\bar{\alpha}_{i-1}}(1-\bar{\alpha}_{t-1})}{\sqrt{\bar{\alpha}_{t-1}}(1-\bar{\alpha}_{i-1})}\right]d_{\theta, i}. \nonumber
\end{align}
At the end of the trajectory, the estimation gap is  
\begin{equation*}
\nonumber
    d_{\theta} = \bar{d}_{\theta, 2}
    = \sum_{i=3}^{T}\left[ \frac{\sqrt{\bar{\alpha}_{i-1}}(1-\bar{\alpha}_{1})}{\sqrt{\alpha_{1}}(1-\bar{\alpha}_{i-1})} \right] d_{\theta, i} + d_{\theta, 2}.
\end{equation*}
The proof is completed by using (\ref{eq:gap}). $\boxdot$

The term $d_{\theta}(\boldsymbol{x}_{0})$ can be considered as the  estimation gap of the model for each example $\boldsymbol{x}_{0}$, while $\mathcal{L}_{\theta}$ represents the overall \textit{estimation error} which is critical for the training process. In typical DPMs, the training process  is often performed by minimizing the conventional square loss  $\mathcal{L}_{simple, t} = \|\boldsymbol{f}_{\theta}(\boldsymbol{x}_{t}, t) - \boldsymbol{\epsilon}_{t}\|^{2}$ at each step $t$, which  may not necessarily minimize $\mathcal{L}_{\theta}$. It means that minimizing $\mathcal{L}_{simple}$ can produce multiple small gaps $d_{\theta, t}$. In the worst case, those small gaps can lead to a non-trivial total gap $d_{\theta}$ as visualized by a 1-D example in Figure~\ref{fig:sampling_trajectory}.
Therefore, a better way to train a DPM is to directly minimize the total gap $d_{\theta}$, instead of trying to minimize each independent term $\mathcal{L}_{simple, t}$. That scenario can be intuitively illustrated in Figure~\ref{fig:expected_sampling_trajectory}. 

Minimizing directly the whole $d_{\theta}$ is challenging due to the requirement of a large number of timesteps, which often leads to a significant memory and computation capability in the training phase. From that observation, we propose a new training loss that aims to minimize the gap term in a slice of trajectory. We name it \textit{sequence-aware loss} based on the idea of considering the error amount of surrounding timesteps. In the next section, we introduce the new training loss and the training algorithm. We also theoretically show that any variants (based on the number of consecutive steps) of that loss function are a tighter upper bound of the estimation error compared to the conventional loss. Finally, we employ that loss function in multiple DPM frameworks and demonstrate its effectiveness on image generation quality.

\subsection{Sequence-aware Training}

Minimizing the mean squared error $||\boldsymbol{f}_{\theta}(\boldsymbol{x}_{t}, t) - \boldsymbol{\epsilon}_{t}||^{2}$ may lead to small gap value at each timestep. However, one critical issue of this approach is that it ignores the relationship between timesteps, which may cause a large total gap $d_{\theta}$ at the end of the trajectory. Instead of optimizing each individual term, minimizing the $d_{\theta}$ should guarantee a good approximation $p_{\theta}(\boldsymbol{x}_{0}|\boldsymbol{x}_{T})$ of the distribution $q(\boldsymbol{x}_{0}|\boldsymbol{x}_{T})$. Nevertheless, that approach often requires a large amount of computation and memory. To address that issue, we propose to minimize the local gap that connects $K$ consecutive steps (for $K>1$):
\begin{align*}
    d_{\theta, t}^{K} &= \sum_{s=t}^{t+K-1} \tau_{s}(\boldsymbol{f}_{\theta}(\boldsymbol{x}_{s}, s) - \boldsymbol{\epsilon}_{s}).
\end{align*}

The \textit{sequence-aware (SA) loss} function for training is:   
\begin{align*}
    \mathcal{L}_{sa} = \mathbf{E}_{t, \boldsymbol{x}_{0}, \boldsymbol{\epsilon}_{t: t+K-1}} \left\| \frac{1}{K} \sum_{s=t}^{t+K-1} \tau_{s}(\boldsymbol{f}_{\theta}(\boldsymbol{x}_{s}, s) - \boldsymbol{\epsilon}_{s}) \right\|^{2},
\end{align*}
where $t \in \{1-K, ..., T\}$ and  $\tau_{s} = 0$ for any $s \notin \{2,..., T\}$.
This training objective enforces the stability in the chain of $K$ consecutive sampling steps. However, we found that optimizing that function independently makes the training error at each timestep quite large, since this SA loss does not strongly constrain the error at individual steps. Therefore, we suggest optimizing $\mathcal{L}_{sa}$ jointly with $\mathcal{L}_{simple}$ to exploit  their advantages, resulting in the following total loss function for training DPMs:
\begin{align}
    \mathcal{L} = \mathcal{L}_{simple} + \lambda \mathcal{L}_{sa},
\end{align}
where $\lambda \ge 0$ is a hyper-parameter that indicates how much we constrain the sampling trajectory. Optimizing the new loss term involves the direction of error at each step. Algorithm~\ref{algo:sa_training} represents the training procedure. In practice, we can ignore constants $\tau_s$ in $\mathcal{L}_{sa}$ since they are often comparable and empirically do not significantly change sample quality.

\begin{algorithm}[t]
\caption{Sequence-aware training}\label{algo:sa_training}
\begin{algorithmic}
\Require Data distribution $q$, number  of timesteps $T$, the noise predictor $\boldsymbol{f}_{\theta}$, number of consecutive steps $K$, hyper-parameter $\lambda$,  learning rate $\eta$. 
\Repeat
\State $\boldsymbol{x}_{0} \sim q(\boldsymbol{x}_{0})$ 
\State $t \sim \mathrm{Uniform}(\{1, \dots, T\})$ 
\For{$k \in \{0,\dots,K-1\}$}
\State $\boldsymbol{\epsilon}_{t+k} \sim \mathcal{N}(\mathbf{0}, \mathbf{I})$ 
\State $\boldsymbol{x}_{t+k} = \sqrt{\bar{\alpha}_{t+k}}\boldsymbol{x}_{0} + \sqrt{1-\bar{\alpha}_{t+k}}\boldsymbol{\epsilon}_{t+k}$ 
\EndFor
\State $\mathcal{L}_{simple} = \|\boldsymbol{f}_{\theta}(\boldsymbol{x}_{t},t)-\boldsymbol{\epsilon}_t\|^{2}$
\State $\mathcal{L}_{sa} = \frac{1}{K^2} \|\sum_{s=t}^{t+K-1}\tau_{s}(\boldsymbol{f}_\theta(\boldsymbol{x}_{s},s) - \boldsymbol{\epsilon}_{s})\|^2$
\State $\mathcal{L} = \mathcal{L}_{simple} + \lambda\mathcal{L}_{sa}$
\State $\theta \leftarrow \theta - \eta \bigtriangledown_{\theta}\mathcal{L}$
\Until converged
\end{algorithmic}
\end{algorithm}

\subsection{Bounding the Estimation Gap}
We have presented the new loss which incorporates more information of the sequential nature of DPMs. We next theoretically show that this loss is tighter than the vanilla loss.

\begin{theorem} \label{thm-lower-bound}
Let $\boldsymbol{f}_{\theta}(\boldsymbol{x}_{s}, s)$ be any noise predictor with parameter $\theta$. Consider the weighted conventional loss function $\mathcal{L}_{simple}^{\tau} \coloneqq \mathbf{E}_{t, \boldsymbol{x}_{0}, \boldsymbol{\epsilon}_{t}} \left[ \tau_{t}^2 \|\boldsymbol{f}_{\theta}(\boldsymbol{x}_{t}, t) - \boldsymbol{\epsilon}_{t}\|^{2}\right]$, where $\tau_{t}$ is defined in Theorem~\ref{thm-Estimation-gap} and $t \in \{2, ...,T\}$. Then 
\begin{equation}
\frac{T-1}{T+K} \mathcal{L}_{simple}^{\tau} \ge  \mathcal{L}_{sa} \ge \frac{1}{(T+K)^2} \mathcal{L}_{\theta}.
\end{equation}
\end{theorem}

\textit{Proof. } By definition,  $\tau_{s} = 0$ for any $s \notin \{2,..., T\}$. We observe that: \\
$(T-1) \mathcal{L}_{simple}^{\tau}$
\begin{align*}
 =& (T-1) \mathbf{E}_{t \in \{2, ...,T\}, \boldsymbol{x}_{0}, \boldsymbol{\epsilon}_{t}} \left[\tau_{t}^2 \|\boldsymbol{f}_{\theta}(\boldsymbol{x}_{t}, t) - \boldsymbol{\epsilon}_{t}\|^{2}\right] \\ 
=& \sum_{t=2}^{T} \mathbf{E}_{\boldsymbol{x}_{0}, \boldsymbol{\epsilon}_{t}} \left[\tau_{t}^2 \|\boldsymbol{f}_{\theta}(\boldsymbol{x}_{t}, t) - \boldsymbol{\epsilon}_{t}\|^{2}\right] \\ 
=& \sum_{t= 1-K}^{T} \mathbf{E}_{ \boldsymbol{x}_{0}, \boldsymbol{\epsilon}_{t: t+K-1}}\left[\frac{1}{K}\sum_{s=t}^{t+K-1}\tau_{s}^2\left\| \boldsymbol{f}_{\theta}(\boldsymbol{x}_{s}, s) - \boldsymbol{\epsilon}_{s}\right\|^{2}\right]
\end{align*}
Jensen's inequality suggests  that 
\begin{multline*}
\frac{1}{K}\sum_{s}\tau_{s}^2\left\| \boldsymbol{f}_{\theta}(\boldsymbol{x}_{s}, s) - \boldsymbol{\epsilon}_{s}\right\|^{2} \\
\ge \left\| \frac{1}{K}\sum_{s} \tau_{s}(\boldsymbol{f}_{\theta}(\boldsymbol{x}_{s}, s) - \boldsymbol{\epsilon}_{s}) \right\|^{2}.
\end{multline*}
Therefore, we have \\ \\
$(T-1) \mathcal{L}_{simple}^{\tau}$
\begin{align*}
\ge& \sum_{t= 1-K}^{T} \mathbf{E}_{ \boldsymbol{x}_{0}, \boldsymbol{\epsilon}_{t: t+K-1}}\left\|\frac{1}{K}\sum_{s=t}^{t+K-1}\tau_{s}( \boldsymbol{f}_{\theta}(\boldsymbol{x}_{s}, s) - \boldsymbol{\epsilon}_{s})\right\|^2 \\
    =& (T+K) \mathcal{L}_{sa}.
\end{align*}

Similarly, by using Jensen's inequality, we can show that\\ \\
$(T+K) \mathcal{L}_{sa} $
\begin{align*}
=& \sum_{t= 1-K}^{T} \mathbf{E}_{ \boldsymbol{x}_{0}, \boldsymbol{\epsilon}_{t: t+K-1}}\left\|\frac{1}{K}\sum_{s=t}^{t+K-1}\tau_{s}( \boldsymbol{f}_{\theta}(\boldsymbol{x}_{s}, s) - \boldsymbol{\epsilon}_{s})\right\|^2  \\
 \ge& \frac{1}{T+K} \mathbf{E}_{ \boldsymbol{x}_{0}, \boldsymbol{\epsilon}}\left\| \sum_{s = 2}^{T}\tau_{s}( \boldsymbol{f}_{\theta}(\boldsymbol{x}_{s}, s) - \boldsymbol{\epsilon}_{s})\right\|^2  \\
 =&  \frac{1}{T+K} \mathcal{L}_{\theta} 
\end{align*}
completing the proof. $\boxdot$

This theorem provides a comparison between our loss and $\mathcal{L}_{simple}^{\tau}$ which is the weighted conventional loss. Since constants $\tau_i$ naturally come from the model formulation and the commonly used loss $\mathcal{L}_{simple}$ ignores those constants, we use the weighted loss for a fair comparison. By using similar arguments with the above proof, it is easy to show that our loss is still tighter than $\mathcal{L}_{simple}$ even when setting every $\tau_i = 1$. This holds for any $K>1$.

\begin{table}[t]
\centering
\small
\rmfamily
\begin{tabular}{|c|c|cccc|}
\hline
\multirow{3}{*}{\textbf{Dataset}} & \multirow{3}{*}{$\boldsymbol{T}$} & \multicolumn{4}{c|}{\textbf{Method}} \\ \cline{3-6} 
 &  & \multicolumn{2}{c|}{DDPM} & \multicolumn{2}{c|}{DDIM} \\ \cline{3-6} 
 &  & \multicolumn{1}{c}{B} & \multicolumn{1}{c|}{SA} & \multicolumn{1}{c}{B} & SA \\ \hline
 \hline
\multirow{3}{*}{CelebA-HQ} & 10 & 54.19 & \multicolumn{1}{c|}{\textbf{53.23}} & 39.29 & \textbf{37.66} \\
 & 50 & 29.04 & \multicolumn{1}{c|}{\textbf{26.73}} & 23.04 & \textbf{20.20} \\
 & 100 & 22.85 & \multicolumn{1}{c|}{\textbf{20.66}} & 22.19 & \textbf{18.98} \\
\multirow{2}{*}{256 $\times$ 256} & 200 & 18.71 & \multicolumn{1}{c|}{\textbf{16.63}} & 22.52 & \textbf{19.27} \\
 & 1000 & 16.03 & \multicolumn{1}{c|}{\textbf{15.32}} & 24.10 & \textbf{20.11} \\ \hline
\end{tabular}
\caption{FID score ($\downarrow$). The results are reported under different number $T$ of timesteps. Here B and SA denote the baseline and our proposed loss.}
\label{table:celebahq}
\end{table}
\begin{figure}[H]
    \centering
    \includegraphics[width=\linewidth]{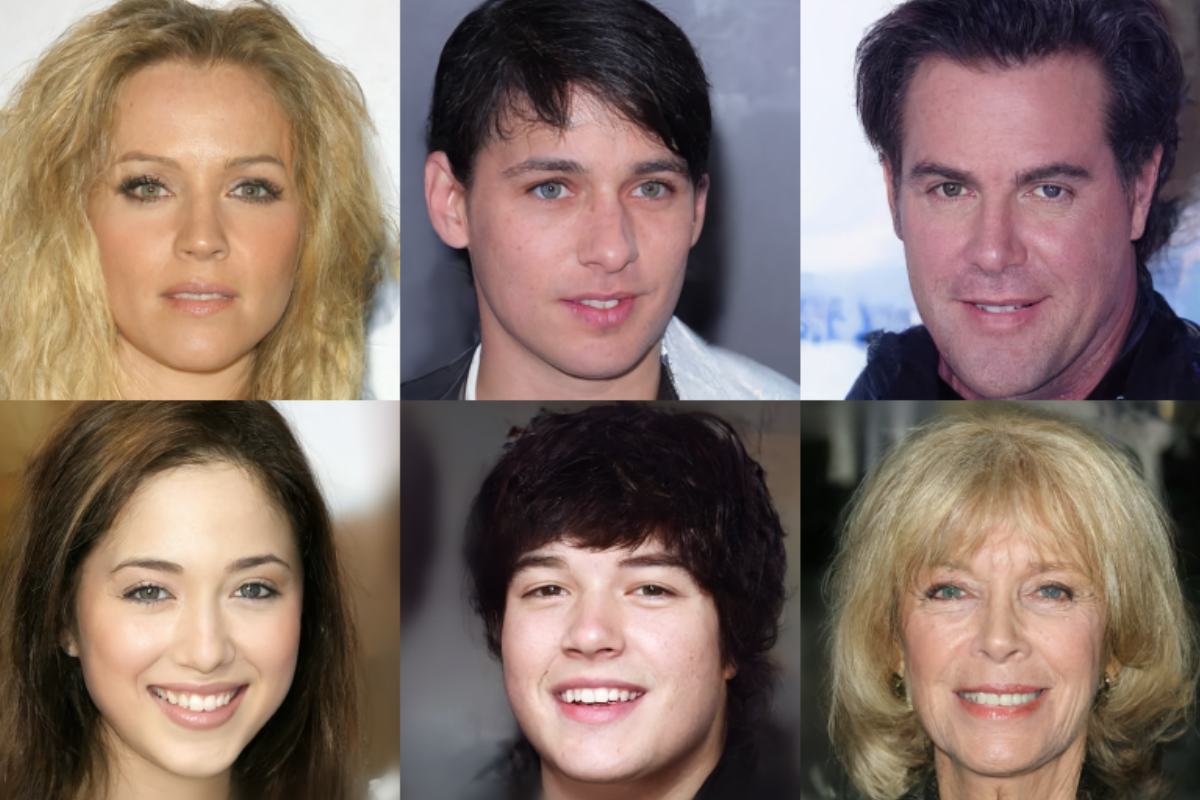}
    \caption{Qualitative results of CelebA-HQ 256 $\times$ 256.}
    \label{fig:celebahq}
\end{figure}
\begin{figure*}[t]
    \centering
    \begin{subfigure}{0.49\textwidth}
        \includegraphics[width=\linewidth]{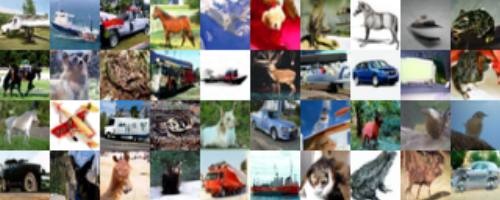}
        \caption{CIFAR10 32 $\times$ 32}
        \label{fig:cifar}
    \end{subfigure}
    \hfill
    \begin{subfigure}{0.49\textwidth}
        \includegraphics[width=\linewidth]{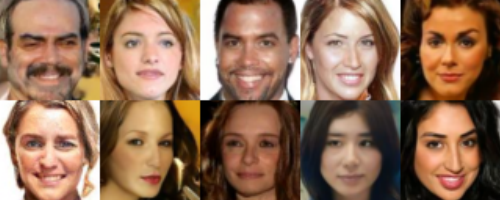}
        \caption{CelebA 64 $\times$ 64}
        \label{fig:celeba}
    \end{subfigure}
    \caption{Qualitative results of (a) CIFAR10 32 $\times$ 32. (b) CelebA 64 $\times$ 64.}
    \label{fig:cifar_celeba}
\end{figure*}
\begin{table*}[!ht]
\centering
\small
\rmfamily
\resizebox{\textwidth}{!}{
\begin{tabular}{|c|c|cccccccc|}
\hline
\multirow{2}{*}{\textbf{Dataset}} & \multirow{2}{*}{\textbf{\# timesteps} $\boldsymbol{T}$} & \multicolumn{8}{c|}{\textbf{Method}} \\ \cline{3-10} 
 &  & B & \multicolumn{1}{c|}{SA} & B+A & \multicolumn{1}{c|}{SA+A} & B+NPR & \multicolumn{1}{c|}{SA+NPR} & B+SN & SA+SN \\ \hline \hline
\multirow{3}{*}{CIFAR10 32$\times$32} & 10 & 41.41 & \multicolumn{1}{c|}{\textbf{30.51}} & 34.19 & \multicolumn{1}{c|}{\textbf{21.66}} & 32.35 & \multicolumn{1}{c|}{\textbf{21.10}} & 24.06 & \textbf{19.53} \\
 & 50 & 15.98 & \multicolumn{1}{c|}{\textbf{9.24}} & 7.20 & \multicolumn{1}{c|}{\textbf{4.20}} & 6.18 & \multicolumn{1}{c|}{\textbf{3.90}} & 4.63 & \textbf{3.61} \\
 & 100 & 11.79 & \multicolumn{1}{c|}{\textbf{6.73}} & 5.31 & \multicolumn{1}{c|}{\textbf{3.43}} & 4.52 & \multicolumn{1}{c|}{\textbf{3.25}} & 3.67 & \textbf{3.10} \\
\multirow{2}{*}{DDPM (LS)} & 200 & 9.15 & \multicolumn{1}{c|}{\textbf{5.47}} & 3.92 & \multicolumn{1}{c|}{\textbf{3.28}} & 3.57 & \multicolumn{1}{c|}{\textbf{3.16}} & 3.31 & \textbf{3.06} \\
 & 1000 & 5.92 & \multicolumn{1}{c|}{\textbf{4.33}} & 3.98 & \multicolumn{1}{c|}{\textbf{3.72}} & 4.10 & \multicolumn{1}{c|}{\textbf{3.84}} & 3.65 & \textbf{3.56} \\ \hline \hline
\multirow{3}{*}{CIFAR10 32$\times$32} & 10 & 34.98 & \multicolumn{1}{c|}{\textbf{24.59}} & 23.41 & \multicolumn{1}{c|}{\textbf{16.66}} & 19.94 & \multicolumn{1}{c|}{\textbf{14.77}} & \textbf{16.33} & 17.23 \\
 & 50 & 11.05 & \multicolumn{1}{c|}{\textbf{6.27}} & 5.42 & \multicolumn{1}{c|}{\textbf{3.78}} & 5.31 & \multicolumn{1}{c|}{\textbf{3.67}} & 4.17 & \textbf{3.97} \\
 & 100 & 8.25 & \multicolumn{1}{c|}{\textbf{4.98}} & 4.45 & \multicolumn{1}{c|}{\textbf{3.53}} & 4.52 & \multicolumn{1}{c|}{\textbf{3.51}} & 3.83 & \textbf{3.64} \\
\multirow{2}{*}{DDPM (CS)} & 200 & 6.69 & \multicolumn{1}{c|}{\textbf{4.40}} & 4.04 & \multicolumn{1}{c|}{\textbf{3.53}} & 4.10 & \multicolumn{1}{c|}{\textbf{3.54}} & 3.72 & \textbf{3.61} \\
 & 1000 & 4.95 & \multicolumn{1}{c|}{\textbf{4.05}} & 4.26 & \multicolumn{1}{c|}{\textbf{3.84}} & 4.27 & \multicolumn{1}{c|}{\textbf{3.87}} & 4.07 & \textbf{3.83} \\ \hline \hline
\multirow{3}{*}{CelebA 64$\times$64} & 10 & 36.69 & \multicolumn{1}{c|}{\textbf{32.15}} & 28.99 & \multicolumn{1}{c|}{\textbf{27.08}} & 28.37 & \multicolumn{1}{c|}{\textbf{26.73}} & \textbf{20.60} & 26.22 \\
 & 50 & 18.96 & \multicolumn{1}{c|}{\textbf{17.59}} & 11.23 & \multicolumn{1}{c|}{\textbf{9.43}} & 10.89 & \multicolumn{1}{c|}{\textbf{9.42}} & 7.88 & \textbf{7.01} \\
 & 100 & 14.31 & \multicolumn{1}{c|}{\textbf{12.77}} & 8.08 & \multicolumn{1}{c|}{\textbf{6.53}} & 8.23 & \multicolumn{1}{c|}{\textbf{6.84}} & 5.89 & \textbf{5.18} \\
\multirow{2}{*}{DDPM} & 200 & 10.48 & \multicolumn{1}{c|}{\textbf{9.14}} & 6.51 & \multicolumn{1}{c|}{\textbf{5.02}} & 7.03 & \multicolumn{1}{c|}{\textbf{5.49}} & 5.02 & \textbf{4.04} \\
 & 1000 & 5.95 & \multicolumn{1}{c|}{\textbf{4.69}} & 5.21 & \multicolumn{1}{c|}{\textbf{3.99}} & 5.33 & \multicolumn{1}{c|}{\textbf{4.00}} & 4.42 & \textbf{3.56} \\ \hline \hline
\multirow{3}{*}{CelebA 64$\times$64} & 10 & 20.54 & \multicolumn{1}{c|}{\textbf{12.88}} & 15.62 & \multicolumn{1}{c|}{\textbf{10.52}} & 14.98 & \multicolumn{1}{c|}{\textbf{10.48}} & \textbf{10.20} & 19.29 \\
 & 50 & 9.33 & \multicolumn{1}{c|}{\textbf{7.01}} & 6.13 & \multicolumn{1}{c|}{\textbf{4.18}} & 6.04 & \multicolumn{1}{c|}{\textbf{4.25}} & 3.83 & \textbf{3.19} \\
 & 100 & 6.60 & \multicolumn{1}{c|}{\textbf{4.81}} & 4.29 & \multicolumn{1}{c|}{\textbf{3.02}} & 4.27 & \multicolumn{1}{c|}{\textbf{3.13}} & 3.04 & \textbf{2.62} \\
\multirow{2}{*}{DDIM} & 200 & 4.96 & \multicolumn{1}{c|}{\textbf{3.69}} & 3.46 & \multicolumn{1}{c|}{\textbf{2.61}} & 3.59 & \multicolumn{1}{c|}{\textbf{2.76}} & 2.85 & \textbf{2.49} \\
 & 1000 & 3.40 & \multicolumn{1}{c|}{\textbf{2.98}} & 3.13 & \multicolumn{1}{c|}{\textbf{2.74}} & 3.15 & \multicolumn{1}{c|}{\textbf{2.78}} & 2.90 & \textbf{2.66} \\ \hline
\end{tabular}
}
\caption{FID score ($\downarrow$). The results are reported under different numbers of timesteps $T$. Here B and SA denote the baseline and our proposed method. A, NPR, and SN denote Analytic-DPM, NPR-DPM, and SN-DPM, respectively.}
\label{table:fid}
\end{table*}

\begin{table*}[!ht]
\centering
\small    
\rmfamily
\resizebox{\textwidth}{!}{
\begin{tabular}{|c|c|cccccccc|}
\hline
\multirow{2}{*}{\textbf{Dataset}} & \multirow{2}{*}{\textbf{\# timesteps} $\boldsymbol{T}$} & \multicolumn{8}{c|}{\textbf{Method}} \\ \cline{3-10} 
 &  & B & \multicolumn{1}{c|}{SA} & B+A & \multicolumn{1}{c|}{SA+A} & B+NPR & \multicolumn{1}{c|}{SA+NPR} & B+SN & SA+SN \\ \hline \hline
\multirow{3}{*}{CIFAR10 32$\times$32} & 10 & 6.93 & \multicolumn{1}{c|}{\textbf{7.55}} & 8.05 & \multicolumn{1}{c|}{\textbf{8.50}} & 8.17 & \multicolumn{1}{c|}{\textbf{8.53}} & 8.10 & \textbf{8.42} \\
 & 50 & 8.34 & \multicolumn{1}{c|}{\textbf{8.82}} & 9.53 & \multicolumn{1}{c|}{\textbf{9.62}} & 9.51 & \multicolumn{1}{c|}{\textbf{9.63}} & 9.49 & \textbf{9.65} \\
 & 100 & 8.59 & \multicolumn{1}{c|}{\textbf{9.04}} & 9.59 & \multicolumn{1}{c|}{\textbf{9.74}} & 9.55 & \multicolumn{1}{c|}{\textbf{9.70}} & 9.47 & \textbf{9.73} \\
\multirow{2}{*}{DDPM (LS)} & 200 & 8.81 & \multicolumn{1}{c|}{\textbf{9.15}} & 9.59 & \multicolumn{1}{c|}{\textbf{9.72}} & 9.49 & \multicolumn{1}{c|}{\textbf{9.62}} & 9.50 & \textbf{9.65} \\
 & 1000 & 9.03 & \multicolumn{1}{c|}{\textbf{9.24}} & 9.17 & \multicolumn{1}{c|}{\textbf{9.37}} & 9.18 & \multicolumn{1}{c|}{\textbf{9.35}} & 9.24 & \textbf{9.41} \\ \hline \hline
\multirow{3}{*}{CIFAR10 32$\times$32} & 10 & 7.48 & \multicolumn{1}{c|}{\textbf{7.97}} & 8.05 & \multicolumn{1}{c|}{\textbf{8.37}} & 8.21 & \multicolumn{1}{c|}{\textbf{8.49}} & 8.47 & \textbf{8.48} \\
 & 50 & 8.53 & \multicolumn{1}{c|}{\textbf{9.09}} & 8.97 & \multicolumn{1}{c|}{\textbf{9.43}} & 9.02 & \multicolumn{1}{c|}{\textbf{9.45}} & 9.10 & \textbf{9.46} \\
 & 100 & 8.71 & \multicolumn{1}{c|}{\textbf{9.20}} & 9.07 & \multicolumn{1}{c|}{\textbf{9.52}} & 9.09 & \multicolumn{1}{c|}{\textbf{9.53}} & 9.16 & \textbf{9.54} \\
\multirow{2}{*}{DDPM (CS)} & 200 & 8.84 & \multicolumn{1}{c|}{\textbf{9.31}} & 9.14 & \multicolumn{1}{c|}{\textbf{9.55}} & 9.15 & \multicolumn{1}{c|}{\textbf{9.54}} & 9.18 & \textbf{9.54} \\
 & 1000 & 8.94 & \multicolumn{1}{c|}{\textbf{9.45}} & 9.04 & \multicolumn{1}{c|}{\textbf{9.52}} & 9.04 & \multicolumn{1}{c|}{\textbf{9.52}} & 9.06 & \textbf{9.54} \\ \hline
\end{tabular}
}
\caption{IS metric ($\uparrow$). The results are reported under different numbers of timesteps $T$. Here B and SA denote the baseline  and our proposed method. A, NPR, and SN denote Analytic-DPM, NPR-DPM, and SN-DPM, respectively.}
\label{table:is}
\end{table*}

\section{Experiments}

\subsection{Image Generation} \label{sec:image_generation}

\subsubsection{Experimental setup:} 
In this experiment, we apply the proposed loss to the vanilla DPM, referred to as SA-$K$-DPM, where $K$ denotes the number of consecutive steps. We evaluate the SA-2-DPM (which we will call SA-DPM for brevity) both individually and in combination with covariance estimation methods, including Analytic-DPM \cite{bao2022analyticdpm}, NPR-DPM and SN-DPM \cite{bao2022estimating}. All settings and hyperparameters are kept unchanged from \cite{song2020denoising}. In particular, the experiments are conducted on: CIFAR10 32$\times$32 \cite{krizhevsky2009learning}, CelebA 64$\times$64 \cite{liu2015faceattributes} and one higher-resolution dataset CelebA-HQ 256$\times$256 \cite{karras2017progressive}. For CIFAR10, the models are trained with two different forward noise schedules: the linear schedule (LS) \cite{ho2020denoising} and the cosine schedule (CS) \cite{nichol2021improved}. The sampling timesteps for all the datasets are set to $\{10, 50, 100, 200, 1000\}$. For the evaluation, we compute the FID between 50$k$ generated images and the pre-computed statistics of the datasets. See more details in Appendix C.1.

\subsubsection{Performance Comparison:} 
The summary of sampling performance for CIFAR10 and CelebA is presented in Table~\ref{table:fid} and \ref{table:is}. Table~\ref{table:celebahq} presents the results for the remaining dataset CelebA-HQ. Evidently,  SA-DPM exhibits a substantial performance improvement over the original DPM, regardless of whether the number of timesteps is small or large. With a large number of timesteps, the original DPM can  fully leverage gradient guidance from the  denoising model across finer sampling iterations to generate higher-quality samples. However, as the number of timesteps is reduced from 1000 down to 10, the performance gains of our SA-DPM become more pronounced. As observed from those tables, for many settings, 50 or 100 timesteps are sufficient for our method to achieve a similar FID level with prior methods which use 1000 timesteps. This suggests a significant advantage of our new loss to improve both training and inference in DPMs. For qualitative results, we provide the generated samples of our SA-DPM in Figure \ref{fig:celebahq} and \ref{fig:cifar_celeba}. 

In addition, we also combine our proposed loss with the three covariance estimation methods (Analytic-DPM, NPR-DPM, and SN-DPM) on two datasets: CIFAR10 and CelebA. Table~\ref{table:fid} and \ref{table:is} show that our loss can boost significantly the image quality. This could be attributed to the capability of our loss to enhance the estimation of the mean of the backward Gaussian distributions in the sampling procedure. So when incorporating the additional covariance estimation methods, the generated image quality is further improved. We further provide synthesized samples in Appendix C.3.

\subsection{Ablation Study on the Weight $\lambda$}

In the previous subsection, we used the SA-2-DPM with the weight $\lambda$ of $\mathcal{L}_{sa}$ set to 1, which resulted in substantial performance improvements when considering small sampling timesteps as compared to the original DPM. Next, we consider the variations in FID scores for CIFAR10 dataset across different configurations of weight $\lambda \in \{0.5, 1, 2\}$ for SA-2-DPM, $\lambda \in \{0.3, 0.6, 1.5\}$ for SA-3-DPM and $\lambda \in \{0.2, 0.4\}$ for SA-4-DPM. In this experiment, the sampling type of DDPM is used for evaluation. As presented in Table \ref{table:ablation_weight}, all the tested SA-$K$-DPM methods yield better results compared to the vanilla DPM. With different numbers of consecutive steps, the weight $\lambda$ plays a crucial role. Specifically, SA-2-DPM ($\lambda = 1$), SA-3-DPM ($\lambda = 0.3$), and SA-4-DPM ($\lambda = 0.2$) consistently outperform DPM for all numbers of sampling timesteps. However, when the weight $\lambda$ is set much higher,  the quality of generated images will degrade slightly when using a large number of timesteps (e.g., 1000), even though it will be significantly better when using a small number of timesteps.

\begin{table}[t]
\centering
\small
\rmfamily
\scalebox{0.97}{
\begin{tabular}{cllllll}
\hline
\multirow{2}{*}{\textbf{Method}}   & \multicolumn{1}{c}{\multirow{2}{*}{$\boldsymbol{\lambda}$}} & \multicolumn{5}{c}{\textbf{\# timesteps} $\boldsymbol{T}$} \\ \cline{3-7}
 & \multicolumn{1}{c}{} & \multicolumn{1}{c}{10} & \multicolumn{1}{c}{50} & \multicolumn{1}{c}{100} & \multicolumn{1}{c}{200} & \multicolumn{1}{c}{1000} \\ \hline
\multirow{1}{*}{DDPM} & 0 & 41.41 & 15.98 & 11.79 & 9.15 & 5.92 \\ \hline
\multirow{3}{*}{SA-2-DPM} & 0.5 & 35.39 & 12.09 & 8.52 & 6.56 & 5.25 \\
                          & 1.0  & 30.51 & 9.24 & 6.73 & 5.47 & 4.33 \\
                          & 2.0  & 19.14 & 10.59 & 11.21 & 12.34 & 14.20 \\ \hline
\multirow{3}{*}{SA-3-DPM} & 0.3  & 30.49 & 10.27 & 7.63 & 6.44 & 5.47 \\
                          & 0.6  & 23.71 & 9.07  & 7.96 & 7.77 & 8.06 \\
                          & 1.5  & 15.59 & 11.76 & 13.90 & 16.34 & 19.49 \\ \hline 
\multirow{2}{*}{SA-4-DPM} & 0.2  & 32.93 & 10.78 & 7.78 & 6.17 & 4.73 \\
                          & 0.4  & 26.68 & 9.33 & 7.53 & 7.00 & 6.95 \\ \hline
\end{tabular}
}
\caption{FID of CIFAR10 dataset under different weight $\lambda$ of $\mathcal{L}_{sa}$. We use the sampling type of DDPM to synthesize.}
\label{table:ablation_weight}
\end{table}

\subsection{Evaluation on the Estimation Gap}

\begin{figure}[t]
    \centering
    \includegraphics[width=0.9\linewidth]{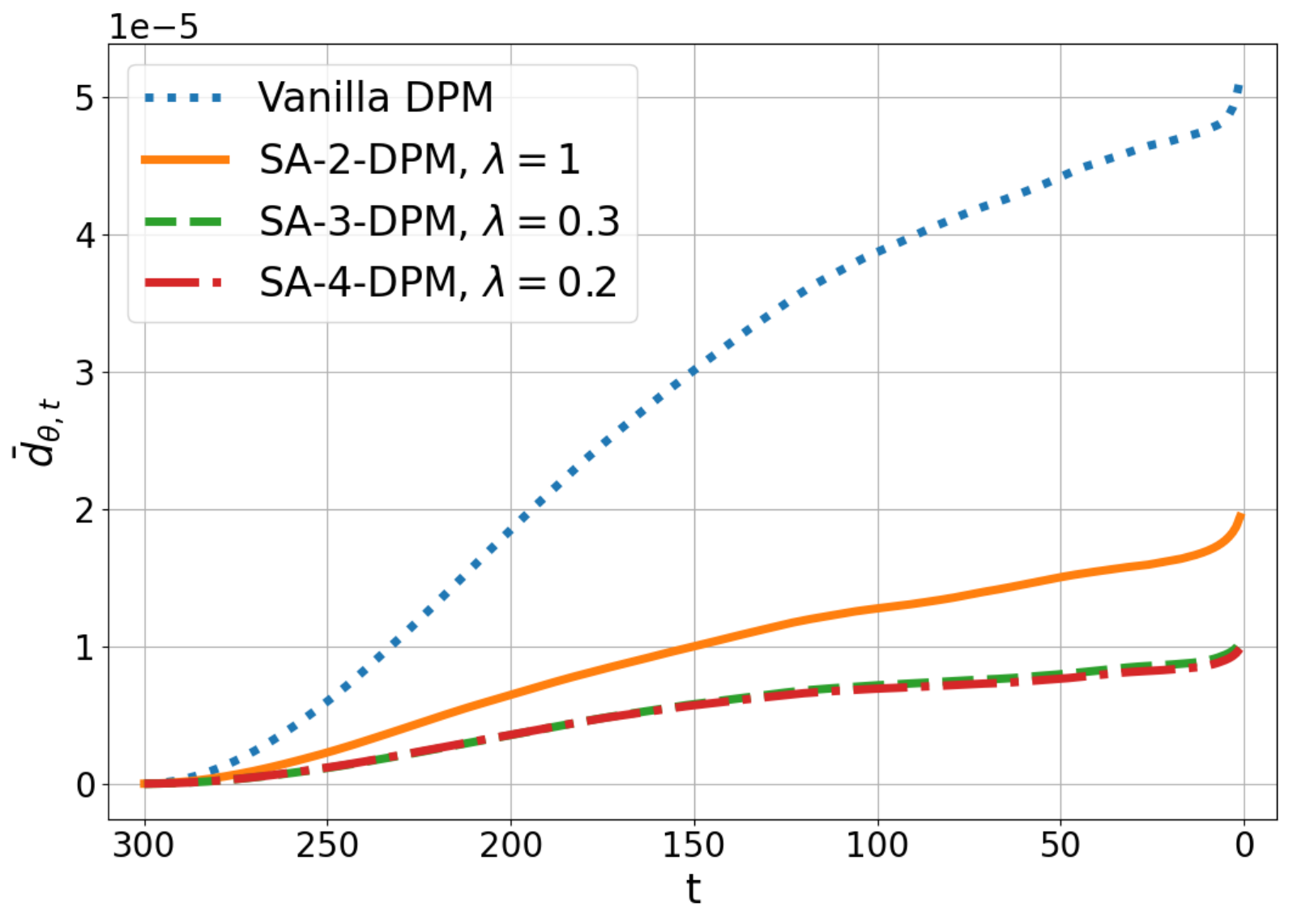}
    \caption{Total gap term $\bar{d}_{\theta, t}$ when sampling image starting from $\boldsymbol{x}_{300}$ on CIFAR10 dataset.}
    \label{fig:gap_loss_300}
\end{figure}
In this experiment, we evaluate the total gap term $\bar{d}_{\theta, t}$ of each trained model during sampling. Because $\bar{d}_{\theta, t}$ contains the weighted sum of the difference between the noise target $\boldsymbol{f}_{\theta}(\boldsymbol{x}_{t}, t)$ and the actual noise $\boldsymbol{\epsilon}_t$, however, during the sampling process starting from Gaussian noise $\boldsymbol{x}_{T} \sim \mathcal{N}(\mathbf{0}, \mathbf{I})$, we cannot know the actual noise due to the unknown input image $\boldsymbol{x}_0$. Therefore, to assess the quantity $\bar{d}_{\theta, t}$ effectively, we take around 2000 input images from the dataset and add noise to them up to time $t = 300$ in order to avoid completely destroying $\boldsymbol{x}_0$. Then, these images $\boldsymbol{x}_{300}$ are used as starting points for the denoising process. At each time step $t$, we calculate the noise target using the formula $\boldsymbol{\epsilon}_t = \frac{\boldsymbol{x}_{t}-\sqrt{\bar{\alpha}_{t}}\boldsymbol{x}_{0}}{\sqrt{1-\bar{\alpha}_{t}}}$, and then we can compute the  gap $\bar{d}_{\theta, t}$. 

Figure \ref{fig:gap_loss_300} illustrates $\bar{d}_{\theta, t}$ of the sampling process of four trained models on CIFAR10 dataset: vanilla DPM, SA-2-DPM, SA-3-DPM and SA-4-DPM. It can be observed that when training with more consecutive timesteps $K$ in $\mathcal{L}_{sa}$, the total gap term is more effectively minimized during the sampling process. Specifically, with SA-2-DPM, at the final timestep of the denoising process, the total gap term is reduced by approximately 2.5 times compared to the base model. We provide more  results in Appendix C.2.

\section{Conclusion}
In this work, we examine the estimation gap between the ground truth and predicted trajectory in the sampling process of DPMs. We then propose a sequence-aware  loss, that optimizes multiple timesteps jointly to leverage their sequential relationship. We theoretically prove that our proposed loss is a tighter upper bound of the estimation gap than the vanilla loss. Our experimental results verify that our loss reduces the estimation gap and enhances the sample quality. Moreover, when combining our loss with advanced techniques, we achieve a significant improvement over the baselines. Therefore, with our new loss, we provide a new benchmark for future research on DPMs. This new loss represents the true loss of a sampling step and therefore may facilitate future deeper understandings of DPMs, such as generalization ability and optimality. One limitation of this work is that our new loss requires the calculation of the network’s output at many timesteps, which makes the training time longer compared to the vanilla loss. 
\section*{Acknowledgements}
This research was partly funded by Vingroup Innovation Foundation (VINIF) under project code VINIF.2022.DA00183.

\bibliography{aaai24}

\begin{thebibliography}{28}
\providecommand{\natexlab}[1]{#1}

\bibitem[{Bao et~al.(2022{\natexlab{a}})Bao, Li, Sun, Zhu, and Zhang}]{bao2022estimating}
Bao, F.; Li, C.; Sun, J.; Zhu, J.; and Zhang, B. 2022{\natexlab{a}}.
\newblock Estimating the Optimal Covariance with Imperfect Mean in Diffusion Probabilistic Models.
\newblock In \emph{International Conference on Machine Learning}, 1555--1584. PMLR.

\bibitem[{Bao et~al.(2022{\natexlab{b}})Bao, Li, Zhu, and Zhang}]{bao2022analyticdpm}
Bao, F.; Li, C.; Zhu, J.; and Zhang, B. 2022{\natexlab{b}}.
\newblock Analytic-{DPM}: an Analytic Estimate of the Optimal Reverse Variance in Diffusion Probabilistic Models.
\newblock In \emph{International Conference on Learning Representations}.

\bibitem[{Chen et~al.(2021)Chen, Zhang, Zen, Weiss, Norouzi, and Chan}]{chen2021wavegrad}
Chen, N.; Zhang, Y.; Zen, H.; Weiss, R.~J.; Norouzi, M.; and Chan, W. 2021.
\newblock WaveGrad: Estimating Gradients for Waveform Generation.
\newblock In \emph{International Conference on Learning Representations}.

\bibitem[{Dinh, Sohl-Dickstein, and Bengio(2017)}]{dinh2017density}
Dinh, L.; Sohl-Dickstein, J.; and Bengio, S. 2017.
\newblock Density estimation using Real {NVP}.
\newblock In \emph{International Conference on Learning Representations}.

\bibitem[{Germain et~al.(2015)Germain, Gregor, Murray, and Larochelle}]{pmlr-v37-germain15}
Germain, M.; Gregor, K.; Murray, I.; and Larochelle, H. 2015.
\newblock MADE: Masked Autoencoder for Distribution Estimation.
\newblock In Bach, F.; and Blei, D., eds., \emph{International Conference on Machine Learning}, volume~37 of \emph{Proceedings of Machine Learning Research}, 881--889. Lille, France: PMLR.

\bibitem[{Ho et~al.(2019)Ho, Chen, Srinivas, Duan, and Abbeel}]{ho2019flow}
Ho, J.; Chen, X.; Srinivas, A.; Duan, Y.; and Abbeel, P. 2019.
\newblock Flow++: Improving flow-based generative models with variational dequantization and architecture design.
\newblock In \emph{International Conference on Machine Learning}, 2722--2730. PMLR.

\bibitem[{Ho, Jain, and Abbeel(2020)}]{ho2020denoising}
Ho, J.; Jain, A.; and Abbeel, P. 2020.
\newblock Denoising diffusion probabilistic models.
\newblock \emph{Advances in neural information processing systems}, 33: 6840--6851.

\bibitem[{Karras et~al.(2018)Karras, Aila, Laine, and Lehtinen}]{karras2017progressive}
Karras, T.; Aila, T.; Laine, S.; and Lehtinen, J. 2018.
\newblock Progressive Growing of GANs for Improved Quality, Stability, and Variation.
\newblock In \emph{International Conference on Learning Representations}.

\bibitem[{Karras et~al.(2022)Karras, Aittala, Aila, and Laine}]{Karras2022edm}
Karras, T.; Aittala, M.; Aila, T.; and Laine, S. 2022.
\newblock Elucidating the Design Space of Diffusion-Based Generative Models.
\newblock In \emph{Advances in Neural Information Processing Systems}.

\bibitem[{Kingma and Dhariwal(2018)}]{kingma2018glow}
Kingma, D.~P.; and Dhariwal, P. 2018.
\newblock Glow: Generative flow with invertible 1x1 convolutions.
\newblock \emph{Advances in Neural Information Processing Systems}, 31.

\bibitem[{Kingma and Welling(2013)}]{kingma2013auto}
Kingma, D.~P.; and Welling, M. 2013.
\newblock Auto-encoding variational bayes.
\newblock \emph{arXiv preprint arXiv:1312.6114}.

\bibitem[{Kong and Ping(2021)}]{kong2021on}
Kong, Z.; and Ping, W. 2021.
\newblock On Fast Sampling of Diffusion Probabilistic Models.
\newblock In \emph{ICML Workshop on Invertible Neural Networks, Normalizing Flows, and Explicit Likelihood Models}.

\bibitem[{Krizhevsky(2012)}]{krizhevsky2009learning}
Krizhevsky, A. 2012.
\newblock Learning Multiple Layers of Features from Tiny Images.
\newblock \emph{University of Toronto}.

\bibitem[{Liu et~al.(2022)Liu, Ren, Lin, and Zhao}]{liu2022pseudo}
Liu, L.; Ren, Y.; Lin, Z.; and Zhao, Z. 2022.
\newblock Pseudo Numerical Methods for Diffusion Models on Manifolds.
\newblock In \emph{International Conference on Learning Representations}.

\bibitem[{Liu et~al.(2015)Liu, Luo, Wang, and Tang}]{liu2015faceattributes}
Liu, Z.; Luo, P.; Wang, X.; and Tang, X. 2015.
\newblock Deep Learning Face Attributes in the Wild.
\newblock In \emph{International Conference on Computer Vision (ICCV)}.

\bibitem[{Nichol and Dhariwal(2021)}]{nichol2021improved}
Nichol, A.~Q.; and Dhariwal, P. 2021.
\newblock Improved denoising diffusion probabilistic models.
\newblock In \emph{International Conference on Machine Learning}, 8162--8171. PMLR.

\bibitem[{Papamakarios, Pavlakou, and Murray(2017)}]{papamakarios2017masked}
Papamakarios, G.; Pavlakou, T.; and Murray, I. 2017.
\newblock Masked autoregressive flow for density estimation.
\newblock \emph{Advances in Neural Information Processing Systems}, 30.

\bibitem[{Rezende, Mohamed, and Wierstra(2014)}]{rezende2014stochastic}
Rezende, D.~J.; Mohamed, S.; and Wierstra, D. 2014.
\newblock Stochastic backpropagation and approximate inference in deep generative models.
\newblock In \emph{International conference on machine learning}, 1278--1286. PMLR.

\bibitem[{Salimans and Ho(2022)}]{salimans2022progressive}
Salimans, T.; and Ho, J. 2022.
\newblock Progressive Distillation for Fast Sampling of Diffusion Models.
\newblock In \emph{International Conference on Learning Representations}.

\bibitem[{Sohl-Dickstein et~al.(2015)Sohl-Dickstein, Weiss, Maheswaranathan, and Ganguli}]{sohl2015deep}
Sohl-Dickstein, J.; Weiss, E.; Maheswaranathan, N.; and Ganguli, S. 2015.
\newblock Deep unsupervised learning using nonequilibrium thermodynamics.
\newblock In \emph{International conference on machine learning}, 2256--2265. PMLR.

\bibitem[{Song, Meng, and Ermon(2021)}]{song2020denoising}
Song, J.; Meng, C.; and Ermon, S. 2021.
\newblock Denoising Diffusion Implicit Models.
\newblock In \emph{International Conference on Learning Representations}.

\bibitem[{Song et~al.(2023)Song, Dhariwal, Chen, and Sutskever}]{song2023consistency}
Song, Y.; Dhariwal, P.; Chen, M.; and Sutskever, I. 2023.
\newblock Consistency models.
\newblock In \emph{International Conference on Machine Learning}.

\bibitem[{Song and Ermon(2019)}]{song2019generative}
Song, Y.; and Ermon, S. 2019.
\newblock Generative modeling by estimating gradients of the data distribution.
\newblock \emph{Advances in neural information processing systems}, 32.

\bibitem[{Song et~al.(2021)Song, Sohl-Dickstein, Kingma, Kumar, Ermon, and Poole}]{song2021scorebased}
Song, Y.; Sohl-Dickstein, J.; Kingma, D.~P.; Kumar, A.; Ermon, S.; and Poole, B. 2021.
\newblock Score-Based Generative Modeling through Stochastic Differential Equations.
\newblock In \emph{International Conference on Learning Representations}.

\bibitem[{Van~den Oord et~al.(2016)Van~den Oord, Kalchbrenner, Espeholt, Vinyals, Graves et~al.}]{van2016conditional}
Van~den Oord, A.; Kalchbrenner, N.; Espeholt, L.; Vinyals, O.; Graves, A.; et~al. 2016.
\newblock Conditional image generation with pixelcnn decoders.
\newblock \emph{Advances in Neural Information Processing Systems}, 29.

\bibitem[{Watson et~al.(2022)Watson, Chan, Ho, and Norouzi}]{watson2022learning}
Watson, D.; Chan, W.; Ho, J.; and Norouzi, M. 2022.
\newblock Learning Fast Samplers for Diffusion Models by Differentiating Through Sample Quality.
\newblock In \emph{International Conference on Learning Representations}.

\bibitem[{Watson et~al.(2021)Watson, Ho, Norouzi, and Chan}]{watson2021learning}
Watson, D.; Ho, J.; Norouzi, M.; and Chan, W. 2021.
\newblock Learning to efficiently sample from diffusion probabilistic models.
\newblock \emph{arXiv preprint arXiv:2106.03802}.

\bibitem[{Zhang, Niwa, and Kleijn(2023)}]{Zhang_2023_CVPR}
Zhang, G.; Niwa, K.; and Kleijn, W.~B. 2023.
\newblock Lookahead Diffusion Probabilistic Models for Refining Mean Estimation.
\newblock In \emph{IEEE/CVF Conference on Computer Vision and Pattern Recognition (CVPR)}, 1421--1429.

\end{thebibliography}
\newpage
\appendix
\section{Further derivation on reverse distribution}
\label{appendix:derivation_on_reverse_distribution}
In this section, we investigate the reverse distribution $q(\boldsymbol{x}_{t}|\boldsymbol{x}_{T}, \boldsymbol{x}_{0})$.

\begin{lemma}
Let $q(\boldsymbol{x}_{0:T})$ be the Markovian forward process where the transition distribution is defined as $q(\boldsymbol{x}_{t}|\boldsymbol{x}_{t-1}) \coloneqq \mathcal{N}(\boldsymbol{x}_{t};\sqrt{1-\beta_{t}}\boldsymbol{x}_{t-1}, \beta_{t}\mathbf{I})$. We have:
\begin{align*}
     q(\boldsymbol{x}_{t-1}|\boldsymbol{x}_{T}, \boldsymbol{x}_{0}) &\coloneqq \mathcal{N}(\boldsymbol{x}_{t-1};\boldsymbol{\mu}^{'}_{t}, \beta^{'}_{t}\mathbf{I}), \\
\end{align*}
where:
\begin{align}
    \boldsymbol{\mu}^{'}_{t} &= \sqrt{\bar{\alpha}_{t-1}}\boldsymbol{x}_{0} + \frac{\sqrt{\bar{\alpha}_{T}}(1-\bar{\alpha}_{t-1})}{\sqrt{\bar{\alpha}_{t-1}(1-\bar{\alpha}_{T})}} \boldsymbol{\epsilon}_{T} \label{eq:init_defined}\\
    \beta^{'}_{T} &= \tilde{\beta}_{T} \nonumber\\
    \beta^{'}_{t} &= \gamma_{2, t}^{2}\beta^{'}_{t+1} + \tilde{\beta}_{t} \nonumber
\end{align}
\end{lemma}

\textit{Proof.} According to (\ref{eq:posterior_mean}), we have:
\begin{align*}
    q(\boldsymbol{x}_{T-1}|\boldsymbol{x}_{T}, \boldsymbol{x}_{0}) = \mathcal{N}(\boldsymbol{x}_{T-1}; \tilde{\boldsymbol{\mu}}_{T}, \tilde{\beta}_{T}\mathbf{I})
\end{align*}
Each sample $\boldsymbol{x}_{T-1} \sim q(\boldsymbol{x}_{T-1}|\boldsymbol{x}_{T}, \boldsymbol{x}_{0})$ can be rewritten as
\begin{eqnarray*}
    \boldsymbol{x}_{T-1} &=& \tilde{\boldsymbol{\mu}}_{T} + \sqrt{\tilde{\beta}_{T}}\tilde{\boldsymbol{\epsilon}}_{T} 
\end{eqnarray*}
and
\begin{eqnarray*}
    \boldsymbol{x}_{T-2} &=& \tilde{\boldsymbol{\mu}}_{T-1} + \sqrt{\tilde{\beta}_{T-1}}\tilde{\boldsymbol{\epsilon}}_{T-1} \\
    &=& \gamma_{1, T-1}\boldsymbol{x}_{0} + \gamma_{2, T-1} \boldsymbol{x}_{T-1} + \sqrt{\tilde{\beta}_{T-1}}\tilde{\boldsymbol{\epsilon}}_{T-1} \\
    &=& \gamma_{1, T-1}\boldsymbol{x}_{0} + \gamma_{2, T-1}\left[ \tilde{\boldsymbol{\mu}}_{T} + \sqrt{\tilde{\beta}_{T}}\tilde{\boldsymbol{\epsilon}}_{T} \right] \\ 
    & &+ \sqrt{\tilde{\beta}_{T-1}}\tilde{\boldsymbol{\epsilon}}_{T-1} \\
    &=& \gamma_{1, T-1}\boldsymbol{x}_{0} + \gamma_{2, T-1}\tilde{\boldsymbol{\mu}}_{T} \\
    & &+ \sqrt{\gamma_{2, T-1}^{2}\tilde{\beta}_{T} +\tilde{\beta}_{T-1}}\boldsymbol{\epsilon}^{'}_{T-1}
\end{eqnarray*}
where $\tilde{\boldsymbol{\epsilon}}_{T-1}, \tilde{\boldsymbol{\epsilon}}_{T}, \boldsymbol{\epsilon}^{'}_{T-1}$ are random noises sampled from $\mathcal{N}(\mathbf{0}, \mathbf{I})$. It suggests that $\boldsymbol{x}_{T-2}$ can be sampled from $\mathcal{N}(\gamma_{1, T-1}\boldsymbol{x}_{0} + \gamma_{2, T-1} \tilde{\boldsymbol{\mu}}_{T}, (\gamma_{2, T-1}^{2}\tilde{\beta}_{T} + \tilde{\beta}_{T-1})\mathbf{I})$. 

Observe further that
\begin{eqnarray*}
    \tilde{\boldsymbol{\mu}}_{T} &=& \gamma_{1, T}\boldsymbol{x}_{0} + \gamma_{2, T}\boldsymbol{x}_{T} \\
    &=& \frac{\sqrt{\bar{\alpha}_{T-1}}\beta_{T}}{1-\bar{\alpha}_{T}}\boldsymbol{x}_{0} + \frac{\sqrt{\alpha_{T}}(1-\bar{\alpha}_{T-1})}{1-\bar{\alpha}_{T}}\boldsymbol{x}_{T} \\
    &=& \frac{\sqrt{\bar{\alpha}_{T-1}}\beta_{T}}{1-\bar{\alpha}_{T}}\boldsymbol{x}_{0} + \frac{\sqrt{\alpha_{T}}(1-\bar{\alpha}_{T-1})}{1-\bar{\alpha}_{T}}(\sqrt{\bar{\alpha}_{T}}\boldsymbol{x}_{0} \\
    & &+ \sqrt{1-\bar{\alpha}_{T}}\boldsymbol{\epsilon}_{T}) \\
    &=& \sqrt{\bar{\alpha}_{T-1}}x_{0} + \frac{\sqrt{\bar{\alpha}_{T}}(1-\bar{\alpha}_{T-1})}{\sqrt{\bar{\alpha}_{T-1}(1-\bar{\alpha}_{T})}}\boldsymbol{\epsilon}_{T}
\end{eqnarray*}
which satisfies equation (\ref{eq:init_defined}) for $T$.

Next, we will use induction. Denote
\begin{align}
    \boldsymbol{\mu}^{'}_{T} &= \tilde{\boldsymbol{\mu}}_{T} \label{eq:mu_T}\\ 
    \boldsymbol{\mu}^{'}_{t} &=  \gamma_{1, t}\boldsymbol{x}_{0} + \gamma_{2, t} \boldsymbol{\mu}^{'}_{t+1} \label{eq:inductive_mu} \\
    \beta^{'}_{T} &= \tilde{\beta}_{T} \nonumber \\
    \beta^{'}_{t} &= \gamma_{2, t}^{2}\beta^{'}_{t+1} + \tilde{\beta}_{t} \nonumber
\end{align}
Assuming $q(\boldsymbol{x}_{t}|\boldsymbol{x}_{T}, \boldsymbol{x}_{0})=\mathcal{N}(\boldsymbol{x}_{t};\boldsymbol{\mu}^{'}_{t+1}, \beta^{'}_{t+1}\mathbf{I})$, we need to prove that $q(\boldsymbol{x}_{t-1}|\boldsymbol{x}_{T}, \boldsymbol{x}_{0}) = \mathcal{N}(\boldsymbol{x}_{t-1};\boldsymbol{\mu}^{'}_{t}, \beta^{'}_{t}\mathbf{I})$. According to (\ref{eq:posterior_mean}):
\begin{align*}
    \boldsymbol{x}_{t-1} &= \tilde{\boldsymbol{\mu}}_{t} + \sqrt{\tilde{\beta}_{t}}\tilde{\boldsymbol{\epsilon}}_{t} \\
    &= \gamma_{1, t}\boldsymbol{x}_{0} + \gamma_{2, t}\boldsymbol{x}_{t} + \sqrt{\tilde{\beta}_{t}}\tilde{\boldsymbol{\epsilon}}_{t} \\
    &= \gamma_{1, t}\boldsymbol{x}_{0} + \gamma_{2, t}\left[ \boldsymbol{\mu}^{'}_{t+1} + \sqrt{\beta^{'}_{t+1}}\boldsymbol{\epsilon}^{'}_{t+1} \right] + \sqrt{\tilde{\beta}_{t}}\tilde{\boldsymbol{\epsilon}}_{t} \\
    &= \left[ \gamma_{1, t} \boldsymbol{x}_{0} + \gamma_{2, t}\boldsymbol{\mu}^{'}_{t+1} \right] + \gamma_{2, t}\sqrt{\beta^{'}_{t+1}}\boldsymbol{\epsilon}^{'}_{t+1} + \sqrt{\tilde{\beta}_{t}}\tilde{\boldsymbol{\epsilon}}_{t} \\
    &= \boldsymbol{\mu}^{'}_{t} + \sqrt{\beta^{'}_{t}}\boldsymbol{\epsilon}^{'}_{t}
\end{align*}
where $\boldsymbol{\epsilon}^{'}_{t}, \boldsymbol{\epsilon}^{'}_{t+1}$ and $\tilde{\boldsymbol{\epsilon}}_{t}$ are sampled from $\mathcal{N}(\mathbf{0}, \mathbf{I})$. As a result, $\boldsymbol{x}_{t-1} \sim \mathcal{N}(\boldsymbol{\mu}^{'}_{t}, \beta^{'}_t \mathbf{I})$.

We next consider $\boldsymbol{\mu}^{'}_{t}$. Note that 
\begin{align*}
    \boldsymbol{\mu}^{'}_{T-1} &= \gamma_{1, T-1}\boldsymbol{x}_{0} + \gamma_{2, T-1}\boldsymbol{\mu}^{'}_{T} \\
    &= \gamma_{1, T-1}\boldsymbol{x}_{0} + \gamma_{2, T-1}\left[ \gamma_{1, T}\boldsymbol{x}_{0} + \gamma_{2, T}\boldsymbol{x}_{T} \right] \\
    &= \left[ \gamma_{1, T-1} + \gamma_{1, T} \gamma_{2, T-1} \right]\boldsymbol{x}_{0} + \gamma_{2, T-1}\gamma_{2, T} \boldsymbol{x}_{T} \\
    &= \frac{\sqrt{\bar{\alpha}_{T-2}}(1-\alpha_{T-1}\alpha_{T})}{1-\bar{\alpha}_{T}}\boldsymbol{x}_{0} + \frac{\sqrt{\bar{\alpha}_{T}}(1-\bar{\alpha}_{T-2})}{\sqrt{\bar{\alpha}_{T-2}}(1-\bar{\alpha}_{T})}\boldsymbol{x}_{T} \\
    &= \sqrt{\bar{\alpha}_{T-2}}\boldsymbol{x}_{0} + \frac{\sqrt{\bar{\alpha}_{T}}(1-\bar{\alpha}_{T-2})}{\sqrt{\bar{\alpha}_{T-2}(1-\bar{\alpha}_{T})}}\boldsymbol{\epsilon}_{T}
\end{align*}
Assuming $\boldsymbol{\mu}^{'}_{t+1} = \sqrt{\bar{\alpha}_{t}}\boldsymbol{x}_{0} + \frac{\sqrt{\bar{\alpha}_{T}}(1-\bar{\alpha}_{t})}{\sqrt{\bar{\alpha}_{t}(1-\bar{\alpha}_{T})}}\boldsymbol{\epsilon}_{T}$, we observe that 
\begin{eqnarray*}
    \boldsymbol{\mu}^{'}_{t} &=& \gamma_{1, t}\boldsymbol{x}_{0} + \gamma_{2, t}\boldsymbol{\mu}^{'}_{t+1} \\
    &=& \left[\frac{\sqrt{\bar{\alpha}_{t-1}}\beta_{t}}{1-\bar{\alpha}_{t}} + \frac{\sqrt{\alpha_{t}}(1-\bar{\alpha}_{t-1})}{1-\bar{\alpha}_{t}}\sqrt{\bar{\alpha}_{t}} \right]\boldsymbol{x}_{0} \\
    & &+ \frac{\sqrt{\bar{\alpha}_{T}}(1-\bar{\alpha}_{t-1})}{\sqrt{\bar{\alpha}_{t-1}(1-\bar{\alpha}_{T})}}\boldsymbol{\epsilon}_{T} \\
    &=& \sqrt{\bar{\alpha}_{t-1}}\boldsymbol{x}_{0} + \frac{\sqrt{\bar{\alpha}_{T}}(1-\bar{\alpha}_{t-1})}{\sqrt{\bar{\alpha}_{t-1}(1-\bar{\alpha}_{T})}}\boldsymbol{\epsilon}_{T}
\end{eqnarray*}
Completing the proof. $\boxdot$

\section{Derivation on Estimation Gap}
\label{appendix:derivation_on_estimation_gap}

Let $\boldsymbol{f}_{\theta}(\boldsymbol{x}_t, t)$ be a noise predictor that will output $\boldsymbol{\epsilon}_{\theta,t} = \boldsymbol{f}_{\theta}(\boldsymbol{x}_t, t)$ for a given input $\boldsymbol{x}_t$ at timestep $t$. This predictor can help us to make prediction $\boldsymbol{x}_{\theta,0}^{(t)}$ for $\boldsymbol{x}_0$ and prediction $\boldsymbol{\tilde{\mu}}_{\theta,t}$ for the mean $\boldsymbol{\tilde{\mu}}_t$.

Denote $d_{\theta, t}$ is the estimation gap at time step $t$: 
\begin{align*}
    d_{\theta, t} = \tilde{\boldsymbol{\mu}}_{\theta, t} - \tilde{\boldsymbol{\mu}}_{t} = \gamma_{1, t}\frac{\sqrt{1-\bar{\alpha}_{t}}}{\sqrt{\bar{\alpha}_{t}}}(\boldsymbol{\epsilon}_{\theta, t} - \boldsymbol{\epsilon}_{t})
\end{align*} 
We have the total gap is 
\begin{align*}
    \bar{d}_{\theta, t} &= \boldsymbol{\mu}^{'}_{\theta, t} - \boldsymbol{\mu}^{'}_{t} \\
    &= d_{\theta, t} + \sum_{i=t+1}^{T}\left[\prod_{s=t}^{i-1}\gamma_{2, s}\right]d_{\theta, i}
\end{align*}

\textit{Proof.} According to (\ref{eq:mu_T}):
\begin{align*}
    \bar{d}_{\theta, T} &= \boldsymbol{\mu}^{'}_{\theta, T} - \boldsymbol{\mu}^{'}_{T} = \tilde{\boldsymbol{\mu}}_{\theta, T} - \tilde{\boldsymbol{\mu}}_{T} = d_{\theta, T}
\end{align*}
Following induction, we assume $\bar{d}_{\theta, t+1} = d_{\theta, t+1} + \sum_{i=t+2}^{T}\left[\prod_{s=t+1}^{i-1}\gamma_{2, s}\right]d_{\theta, i}$. We next need to prove $\bar{d}_{\theta, t} = d_{\theta, t} + \sum_{i=t+1}^{T}\left[\prod_{s=t}^{i-1}\gamma_{2, s}\right]d_{\theta, i}$. Indeed, according to (\ref{eq:inductive_mu}), we have:
\begin{align*}
    \bar{d}_{\theta, t} &= \boldsymbol{\mu}^{'}_{\theta, t} - \boldsymbol{\mu}^{'}_{t} \\
    &= \gamma_{1, t} \boldsymbol{x}_{\theta,0}^{(t)} + \gamma_{2, t} \boldsymbol{\mu}^{'}_{\theta, t+1} - \gamma_{1, t}\boldsymbol{x}_{0} - \gamma_{2, t} \boldsymbol{\mu}^{'}_{t+1} \\
    &= d_{\theta, t} + \gamma_{2, t}\bar{d}_{\theta, t+1} \\
    &= d_{\theta, t} + \gamma_{2, t}\left[ d_{\theta, t+1} + \sum_{i=t+2}^{T}\left[\prod_{s=t+1}^{i-1}\gamma_{2, s}\right]d_{\theta, i} \right] \\
    &= d_{\theta, t} + \sum_{i=t+1}^{T}\left[\prod_{s=t}^{i-1}\gamma_{2, s}\right]d_{\theta, i}
\end{align*}
Completing the proof. $\boxdot$

\section{Experimental Results}
\subsection{Additional Implementation Details}
\label{appendix:addition_implementation_detail}
We implement our sequence-aware diffusion model based on DDIM codebase \url{https://github.com/ermongroup/ddim}. We maintained default hyper-parameter settings across all models to ensure a fair comparison. Specifically, for the CIFAR10 32$\times$32, CelebA 64$\times$64, we use the Adam optimizer with a learning rate of $2 \times 10^{-4}$ and a batch size of 128; for CelebA-HQ 256$\times$256 we use a learning rate of $2 \times 10^{-5}$ and a batch size of 24. The training epoch is 1200 for CIFAR10 and CelebA, and 900 for CelebA-HQ. For all datasets, we use an exponential moving average (EMA) with a rate of 0.9999. We save a checkpoint every 25 epochs and select the one with the best FID on 2000 generated samples for CIFAR10 and 1000 generated samples for other datasets under full 1000 timesteps. After completing the training of the model, the best checkpoint is utilized in combination with the three covariance estimation methods to do inference (Analytic DPM, SN-DPM, and NPR-DPM) by using their official code directly, \url{https://github.com/baofff/Analytic-DPM} and \url{https://github.com/baofff/Extended-Analytic-DPM}. 

For sampling, we use the official implementation of FID \url{https://github.com/mseitzer/pytorch-fid} and Inception Score \url{https://github.com/toshas/torch-fidelity}. We calculate the FID score on 50$k$ generated samples on all datasets. The reference distribution statistics are obtained from \url{https://github.com/NVlabs/denoising-diffusion-gan}. 

We run all experiments using PyTorch 2.0.0 and CUDA 12.1.66 with 1-4 NVIDIA A100 GPUs for each corresponding dataset.

\subsection{Additional estimation gap comparison}
\label{appendix:additional_estimation_gap_comparison}
\begin{figure*}[ht]
    \begin{subfigure}{0.49\textwidth}
        \centering
        \includegraphics[width=\linewidth]{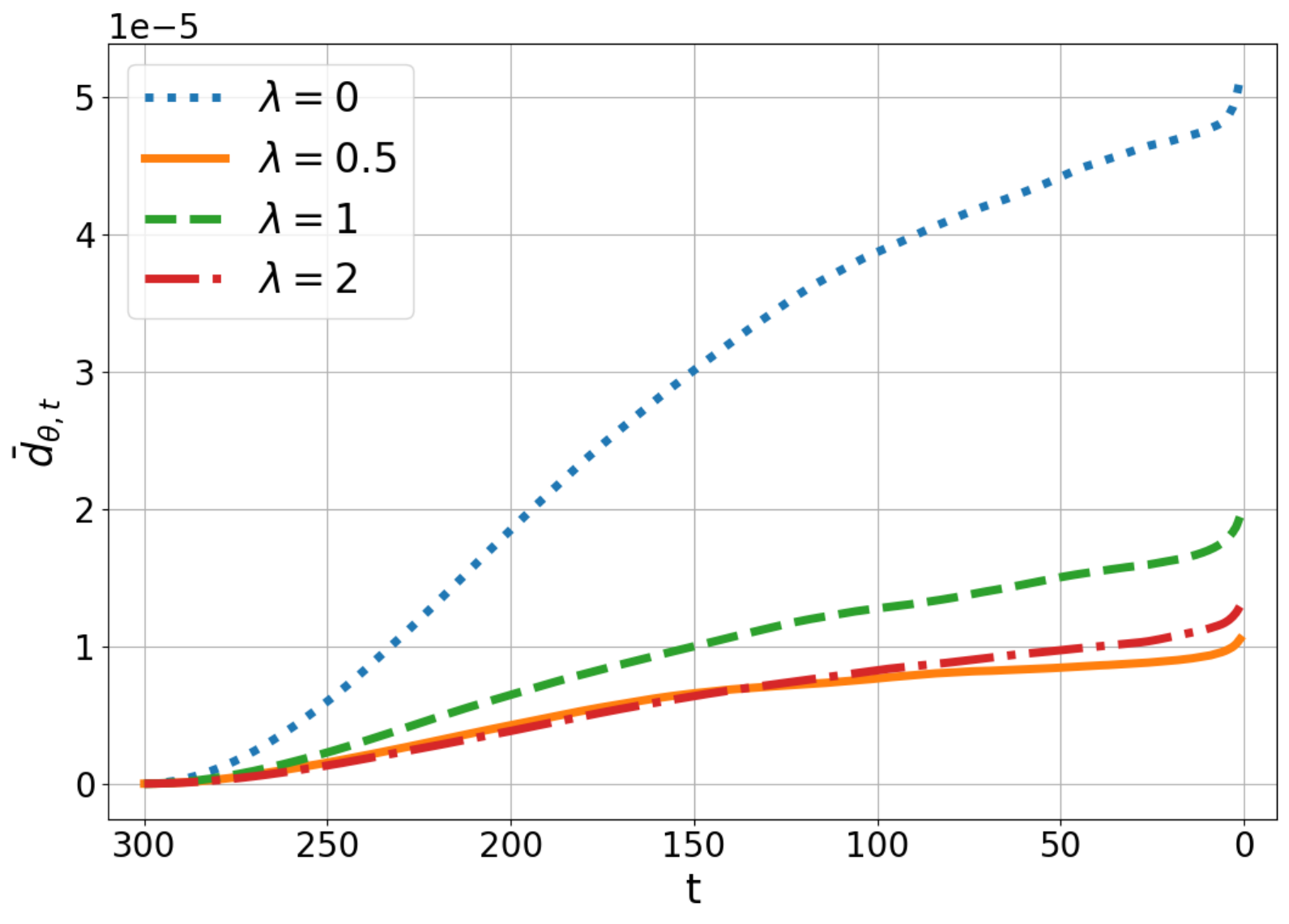}
        \caption{SA-2-DPM (CIFAR10)}
    \end{subfigure}
    \hfill
    \begin{subfigure}{0.49\textwidth}
        \centering
        \includegraphics[width=\linewidth]{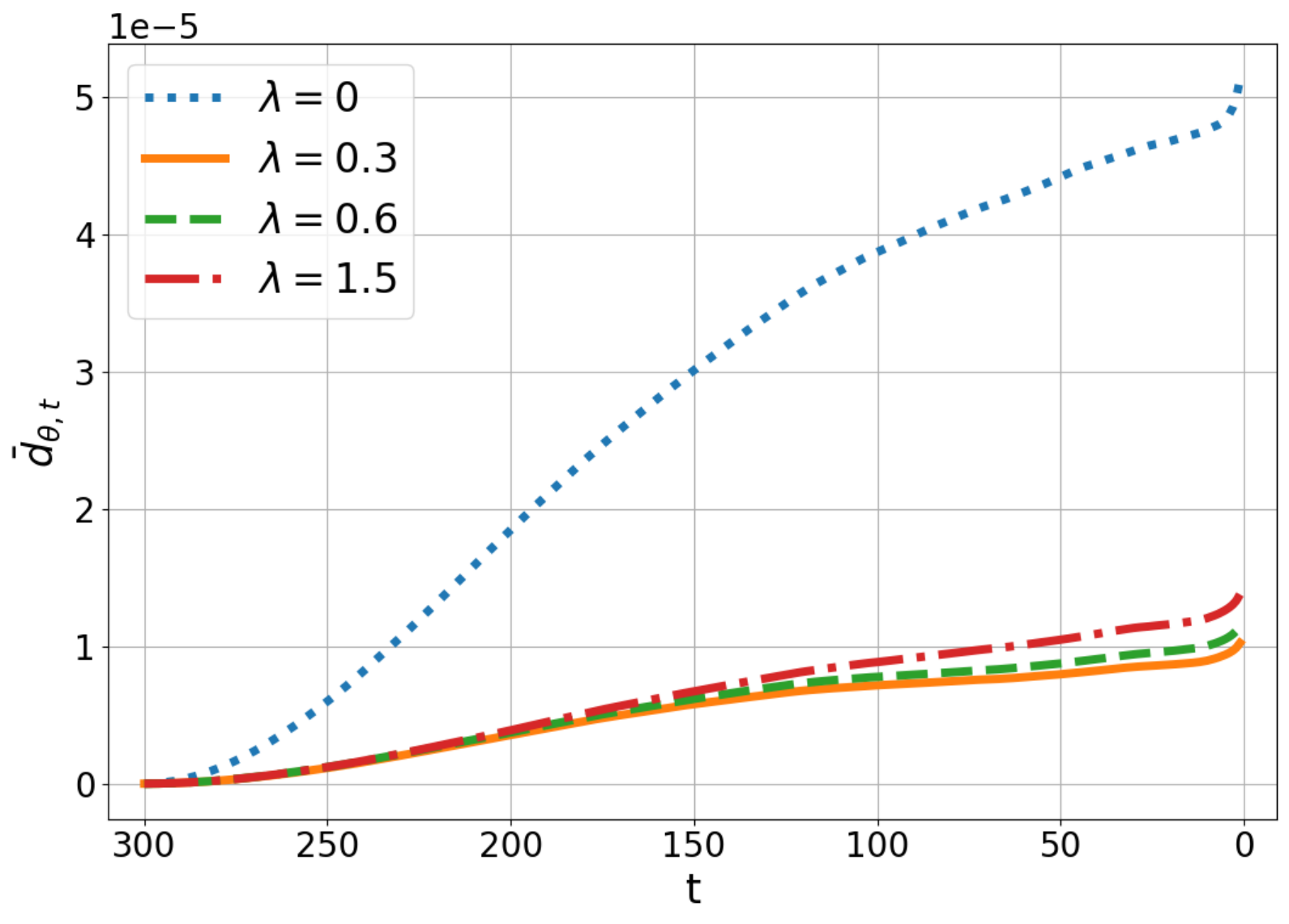}
        \caption{SA-3-DPM (CIFAR10)}
    \end{subfigure}
    
    \vspace{1em} 
    
    \begin{subfigure}{0.49\textwidth}
        \centering
        \includegraphics[width=\linewidth]{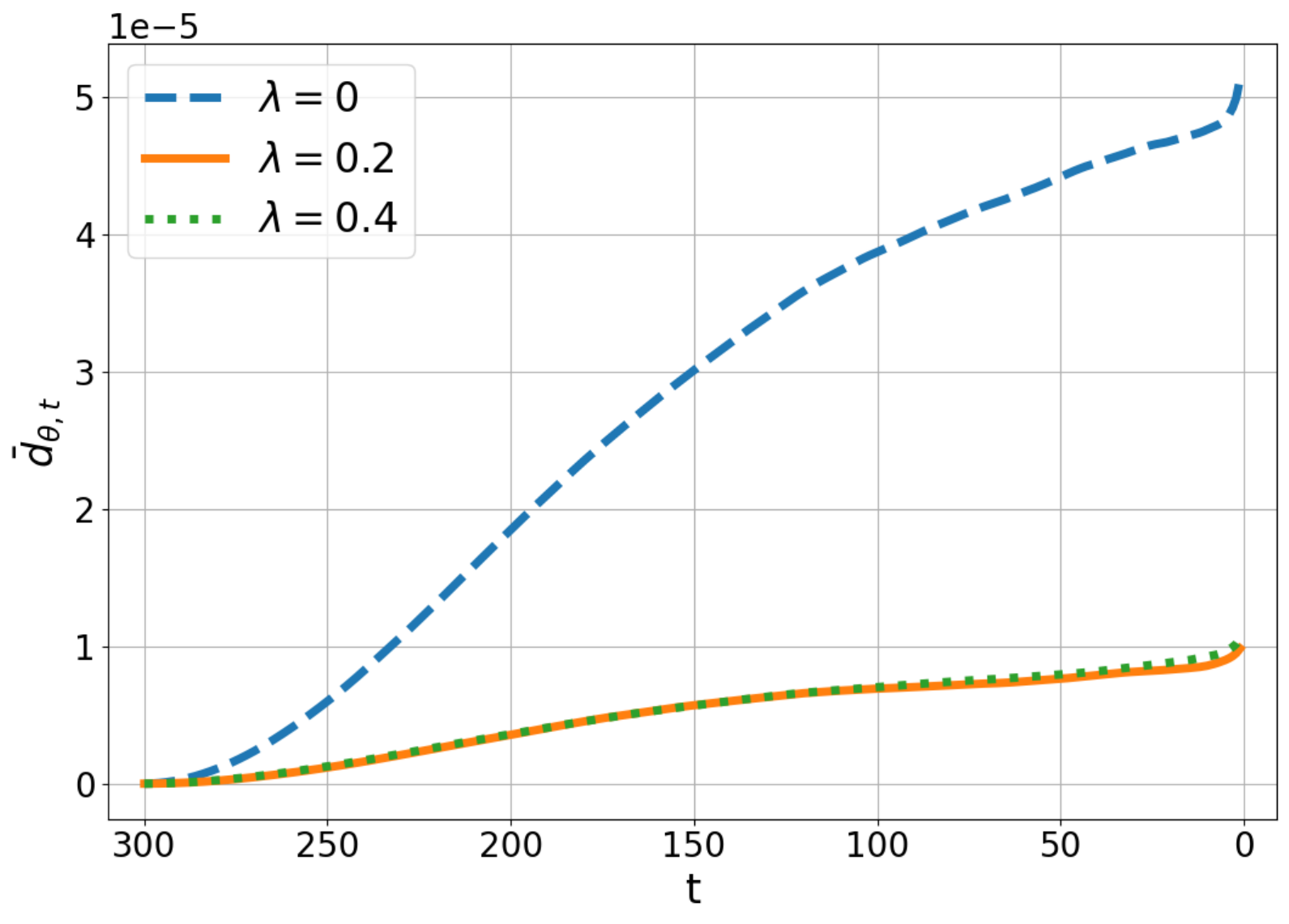}
        \caption{SA-4-DPM (CIFAR10)}
    \end{subfigure}
    \hfill
    \begin{subfigure}{0.49\textwidth}
        \centering
        \includegraphics[width=\linewidth]{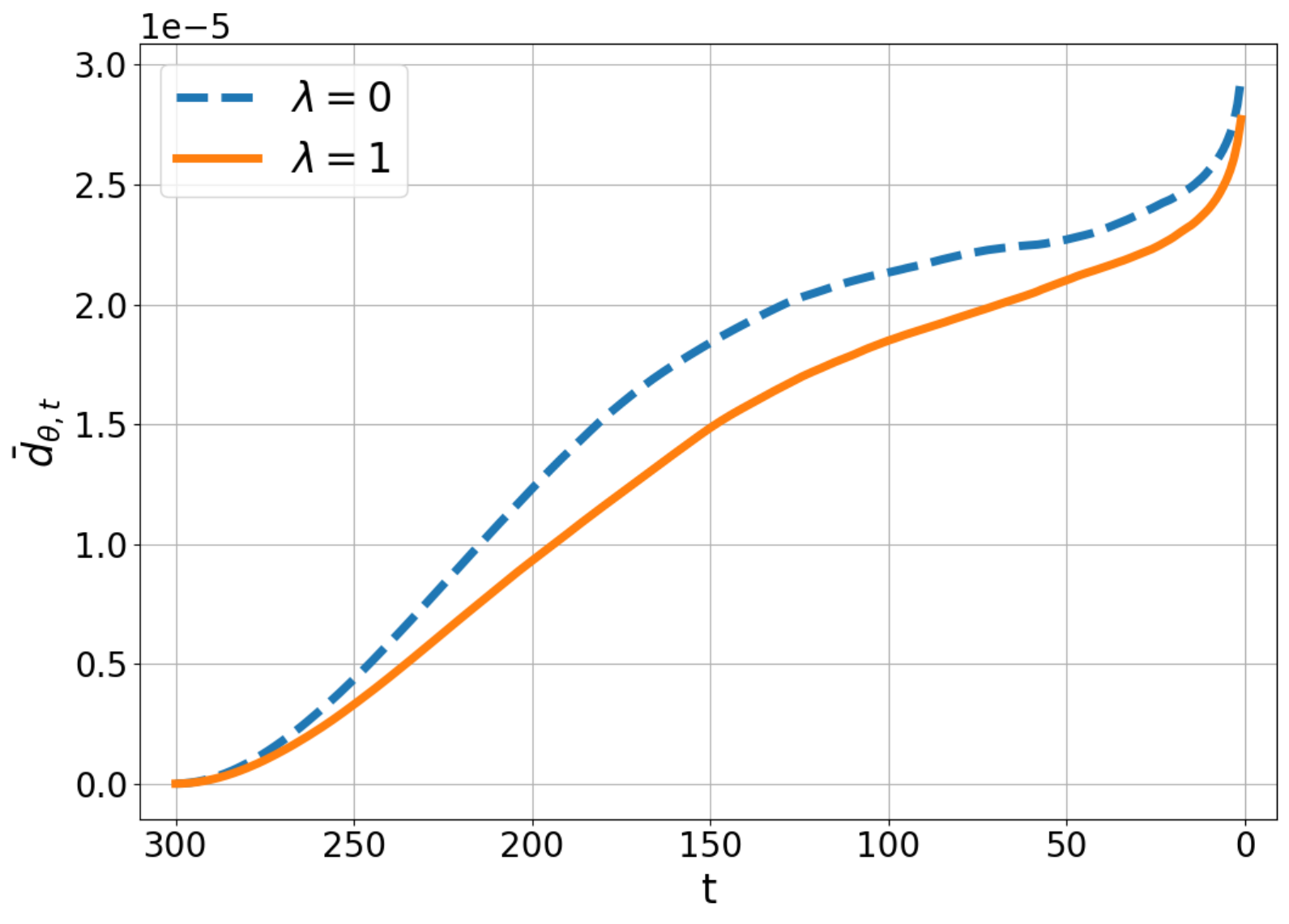}
        \caption{SA-2-DPM (CelebA $64 \times 64$)}
    \end{subfigure}
   
    \caption{Total gap term $\bar{d}_{\theta, t}$ when sampling image starting from $\boldsymbol{x}_{300}$ on CIFAR10 dataset (a, b, c) and CelebA $64 \times 64$ dataset (d). $\lambda = 0$ denotes the Vanilla DPM.}
    \label{fig:appendix_loss_gap}
\end{figure*}

In all settings, the total loss term $d_{\theta}$ of SA-DPM is smaller in comparison with that of the original version. This suggests that the final reverse distribution $q(\boldsymbol{x}_{0}|\boldsymbol{x}_{T})$ is approximated well by $p_{\theta}(\boldsymbol{x}_{0}| \boldsymbol{x}_{T})$. The velocity of $d_{\theta}$ by step is also reduced suggesting the stability in the sampling phase of DPM. We observe that increasing the number of consecutive timesteps stacked in the training phase does not affect much the $d_{\theta}$ and the model performance as presented in Figure~\ref{fig:appendix_loss_gap}. We leave further investigation on the ability of DPM and the effect of long-range training in future works.

\subsection{Samples}
\label{appendix:samples}

\begin{figure*}
    \begin{subfigure}{0.2425\textwidth}
        \centering
        \includegraphics[width=\linewidth]{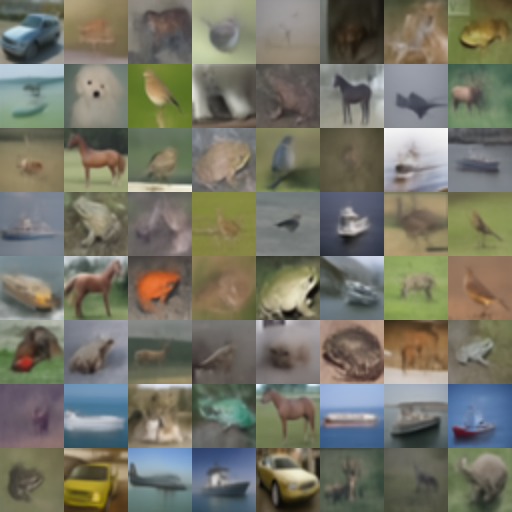}
        \caption{Vanilla-DDPM}
    \end{subfigure}\hspace{0.01\textwidth}%
    \begin{subfigure}{0.2425\textwidth}
        \centering
        \includegraphics[width=\linewidth]{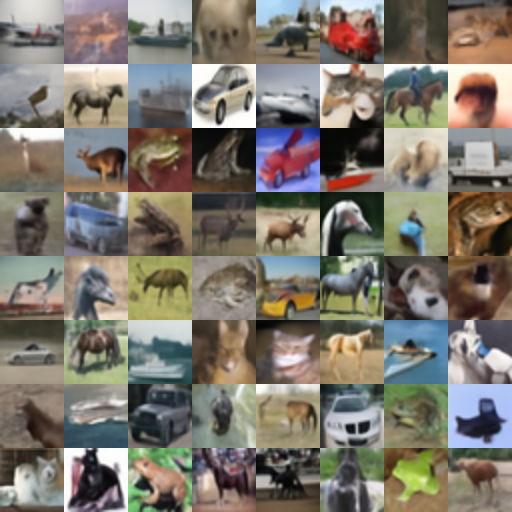}
        \caption{SA-2-DDPM}
    \end{subfigure}\hspace{0.01\textwidth}%
    \begin{subfigure}{0.2425\textwidth}
        \centering
        \includegraphics[width=\linewidth]{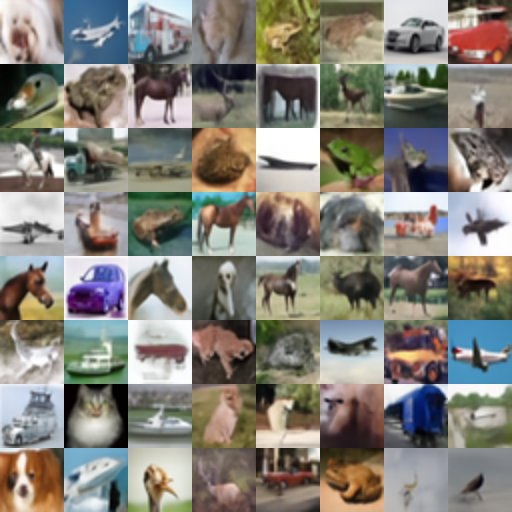}
        \caption{SA-3-DDPM}
    \end{subfigure}\hspace{0.01\textwidth}%
    \begin{subfigure}{0.2425\textwidth}
        \centering
        \includegraphics[width=\linewidth]{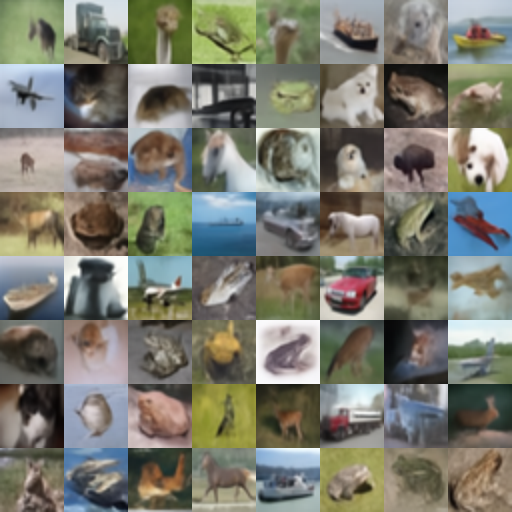}
        \caption{SA-4-DDPM}
    \end{subfigure}
   
    \caption{Generated samples on CIFAR10 ($T=10$).}
    \label{fig:cifar_t10}
\end{figure*}

\begin{figure*}
    \begin{subfigure}{0.2425\textwidth}
        \centering
        \includegraphics[width=\linewidth]{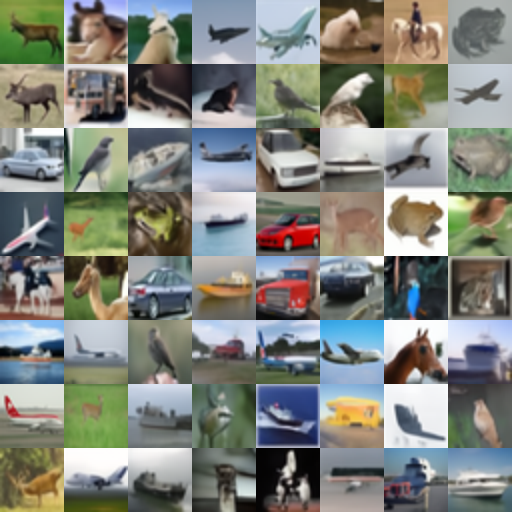}
        \caption{Vanilla-DDPM}
    \end{subfigure}\hspace{0.01\textwidth}%
    \begin{subfigure}{0.2425\textwidth}
        \centering
        \includegraphics[width=\linewidth]{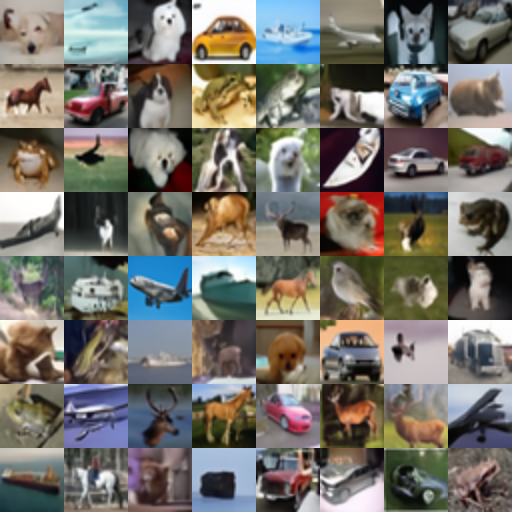}
        \caption{SA-2-DDPM}
    \end{subfigure}\hspace{0.01\textwidth}%
    \begin{subfigure}{0.2425\textwidth}
        \centering
        \includegraphics[width=\linewidth]{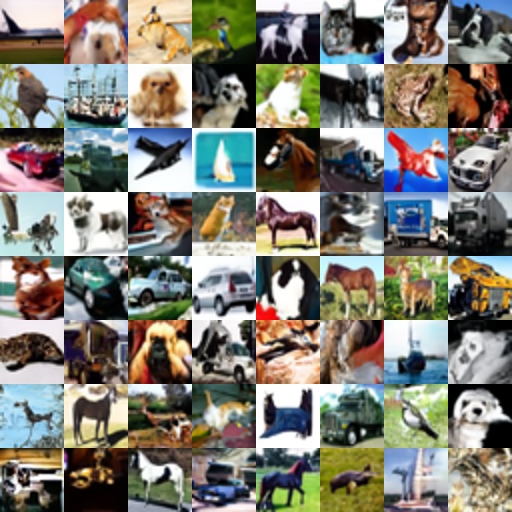}
        \caption{SA-3-DDPM}
    \end{subfigure}\hspace{0.01\textwidth}%
    \begin{subfigure}{0.2425\textwidth}
        \centering
        \includegraphics[width=\linewidth]{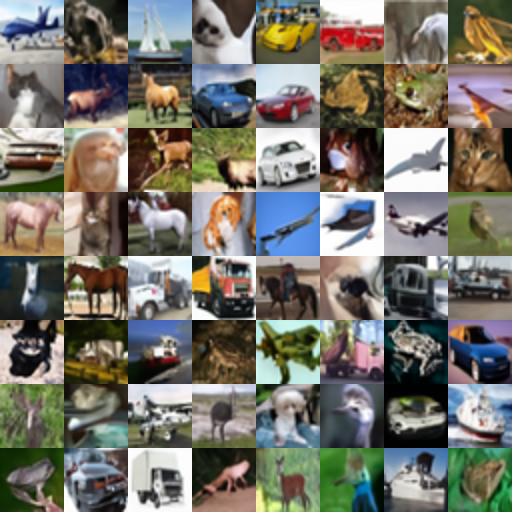}
        \caption{SA-4-DDPM}
    \end{subfigure}
   
    \caption{Generated samples on CIFAR10 ($T=50$).}
    \label{fig:cifar_t50}
\end{figure*}

\begin{figure*}
    \begin{subfigure}{0.2425\textwidth}
        \centering
        \includegraphics[width=\linewidth]{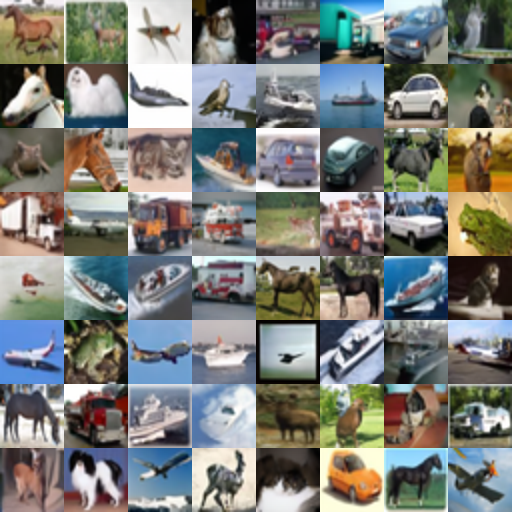}
        \caption{Vanilla-DDPM}
    \end{subfigure}\hspace{0.01\textwidth}%
    \begin{subfigure}{0.2425\textwidth}
        \centering
        \includegraphics[width=\linewidth]{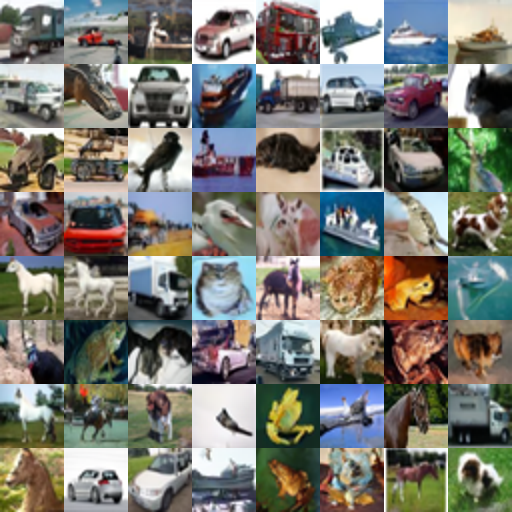}
        \caption{SA-2-DDPM}
    \end{subfigure}\hspace{0.01\textwidth}%
    \begin{subfigure}{0.2425\textwidth}
        \centering
        \includegraphics[width=\linewidth]{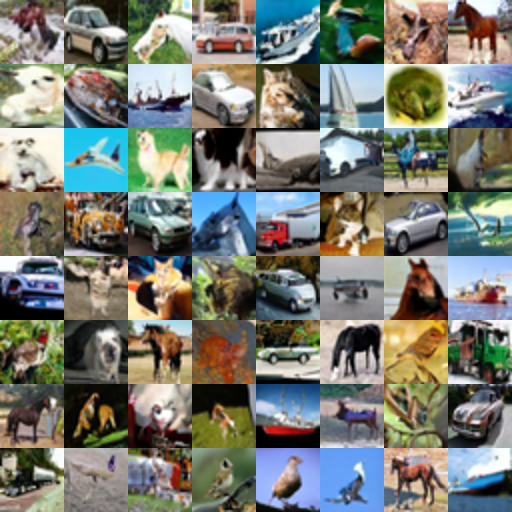}
        \caption{SA-3-DDPM}
    \end{subfigure}\hspace{0.01\textwidth}%
    \begin{subfigure}{0.2425\textwidth}
        \centering
        \includegraphics[width=\linewidth]{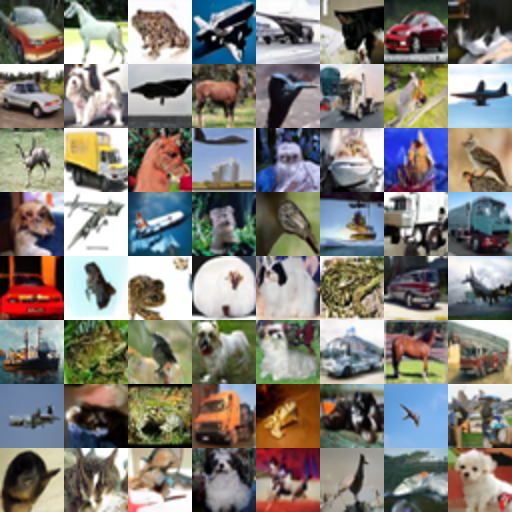}
        \caption{SA-4-DDPM}
    \end{subfigure}
    \caption{Generated samples on CIFAR10 ($T=1000$).}
    \label{fig:cifar_t1000}
\end{figure*}

\begin{figure*}
    \begin{subfigure}{0.2425\textwidth}
        \centering
        \includegraphics[width=\linewidth]{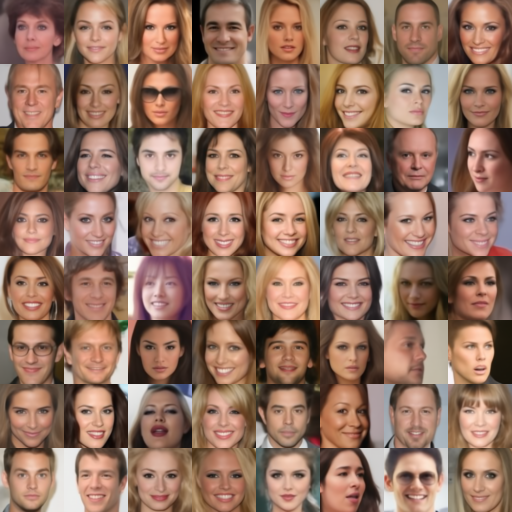}
        \captionsetup{font=scriptsize} 
        \caption{SA-2-DDPM $(T=10)$}
    \end{subfigure}\hspace{0.01\textwidth}%
    \begin{subfigure}{0.2425\textwidth}
        \centering
        \includegraphics[width=\linewidth]{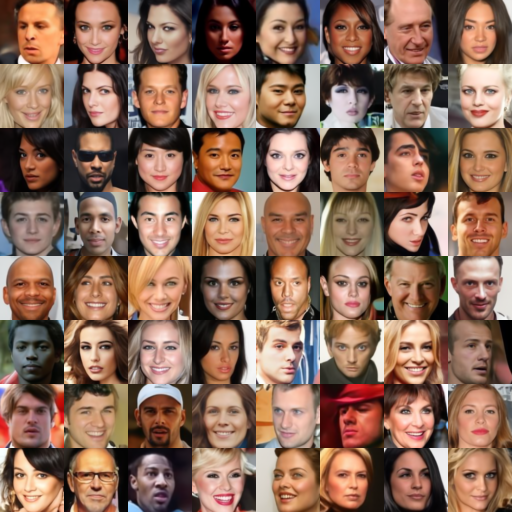}
        \captionsetup{font=scriptsize} 
        \caption{SA-2-DDPM $(T=50)$}
    \end{subfigure}\hspace{0.01\textwidth}%
    \begin{subfigure}{0.2425\textwidth}
        \centering
        \includegraphics[width=\linewidth]{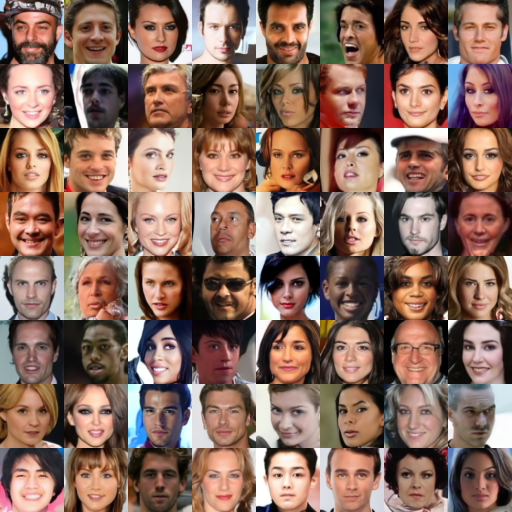}
        \captionsetup{font=scriptsize} 
        \caption{SA-2-DDPM $(T=200)$}
    \end{subfigure}\hspace{0.01\textwidth}%
    \begin{subfigure}{0.2425\textwidth}
        \centering
        \includegraphics[width=\linewidth]{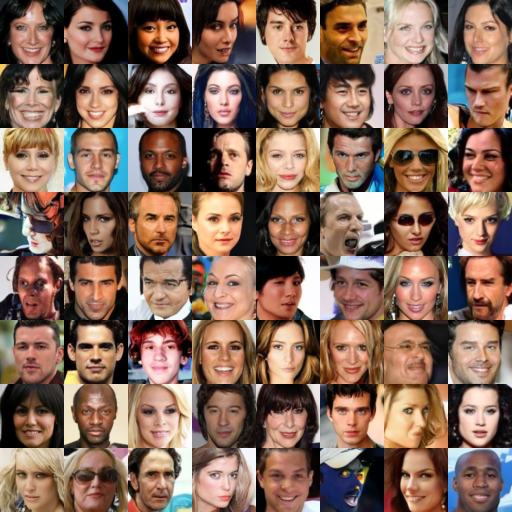}
        \captionsetup{font=scriptsize} 
        \caption{SA-2-DDPM $(T=1000)$}
    \end{subfigure}
\vspace{1em}
    \begin{subfigure}{0.2425\textwidth}
        \centering
        \includegraphics[width=\linewidth]{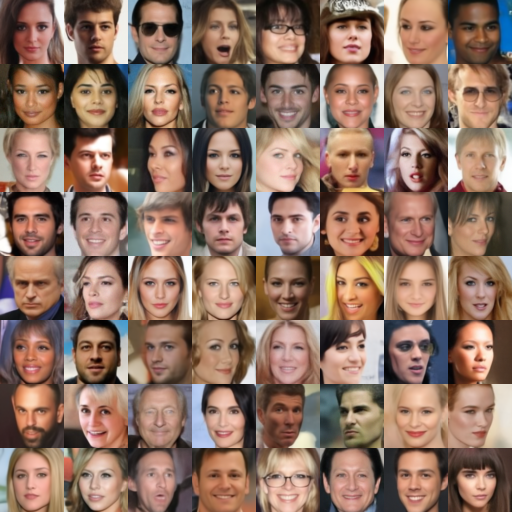}
        \captionsetup{font=scriptsize} 
        \caption{SA-2-DDIM $(T=10)$}
    \end{subfigure}\hspace{0.01\textwidth}%
    \begin{subfigure}{0.2425\textwidth}
        \centering
        \includegraphics[width=\linewidth]{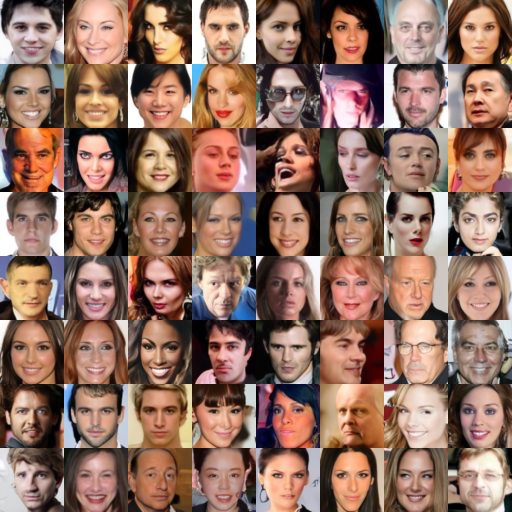}
        \captionsetup{font=scriptsize} 
        \caption{SA-2-DDIM $(T=50)$}
    \end{subfigure}\hspace{0.01\textwidth}%
    \begin{subfigure}{0.2425\textwidth}
        \centering
        \includegraphics[width=\linewidth]{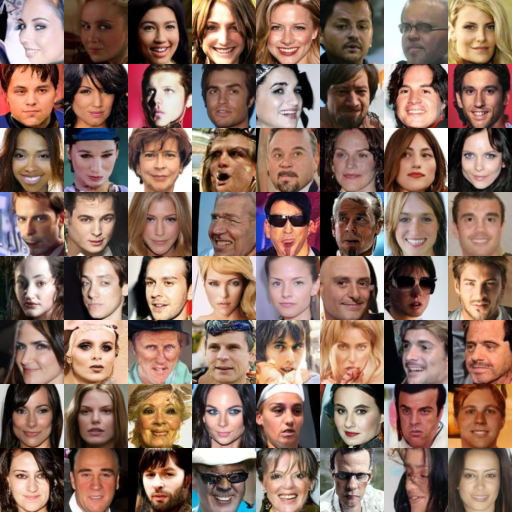}
        \captionsetup{font=scriptsize} 
        \caption{SA-2-DDIM $(T=200)$}
    \end{subfigure}\hspace{0.01\textwidth}%
    \begin{subfigure}{0.2425\textwidth}
        \centering
        \includegraphics[width=\linewidth]{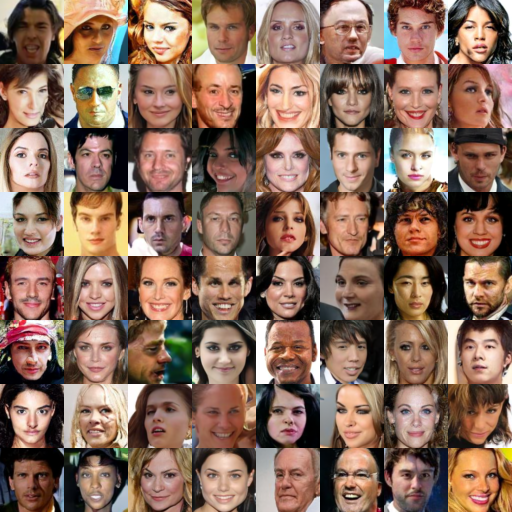}
        \captionsetup{font=scriptsize} 
        \caption{SA-2-DDIM $(T=1000)$}
    \end{subfigure}
   
    \caption{Generated samples on CelebA $64 \times 64$.}
    \label{fig:celeba_sa_2}
\end{figure*}

\begin{figure*}
    \begin{subfigure}{0.48\textwidth}
        \centering
        \includegraphics[width=\linewidth]{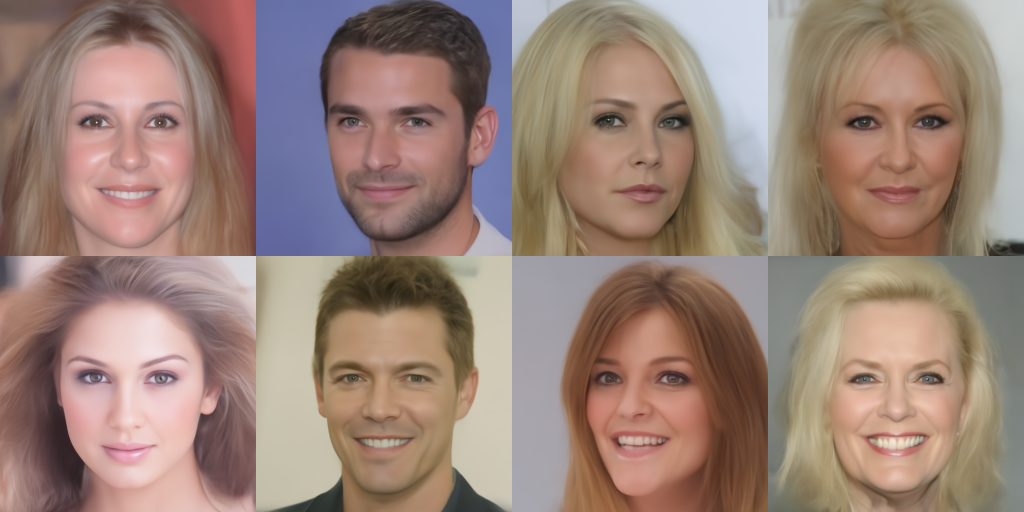}
        \captionsetup{font=scriptsize} 
        \caption{SA-2-DDPM $(T=10)$}
    \end{subfigure}
    \hspace{0.04\textwidth}%
    \begin{subfigure}{0.48\textwidth}
        \centering
        \includegraphics[width=\linewidth]{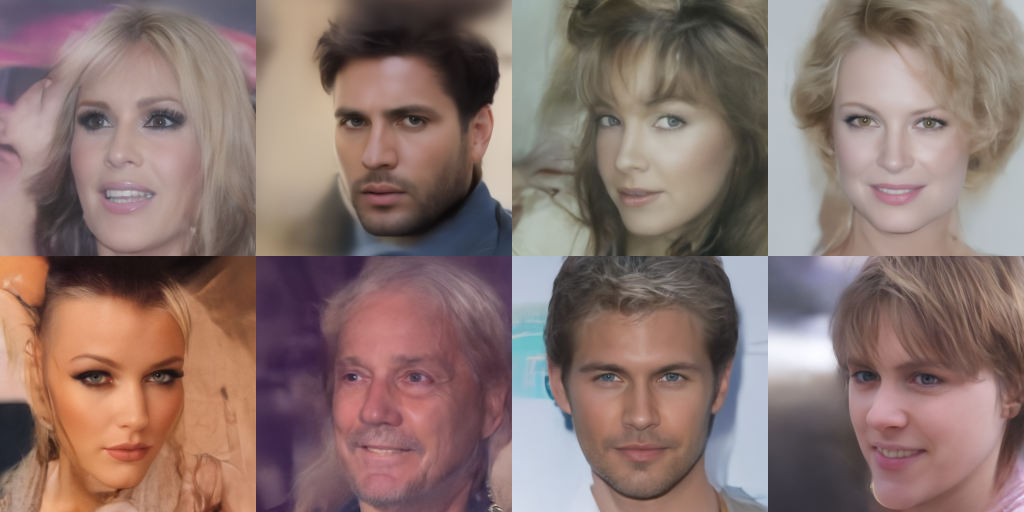}
        \captionsetup{font=scriptsize} 
        \caption{SA-2-DDIM $(T=10)$}
    \end{subfigure}\hspace{0.02\textwidth}%
    \begin{subfigure}{0.48\textwidth}
        \centering
        \includegraphics[width=\linewidth]{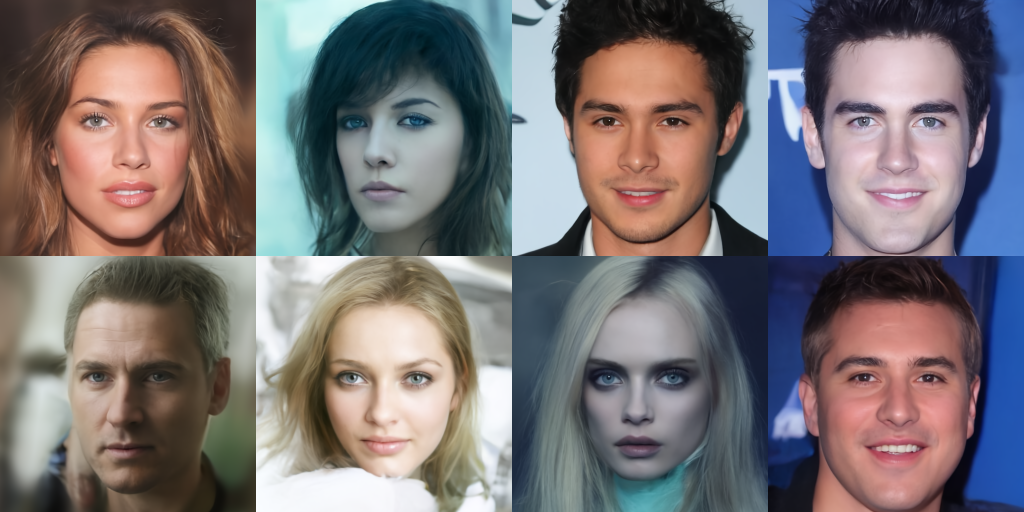}
        \captionsetup{font=scriptsize} 
        \caption{SA-2-DDPM $(T=50)$}
    \end{subfigure}
    \hspace{0.04\textwidth}%
    \begin{subfigure}{0.48\textwidth}
        \centering
        \includegraphics[width=\linewidth]{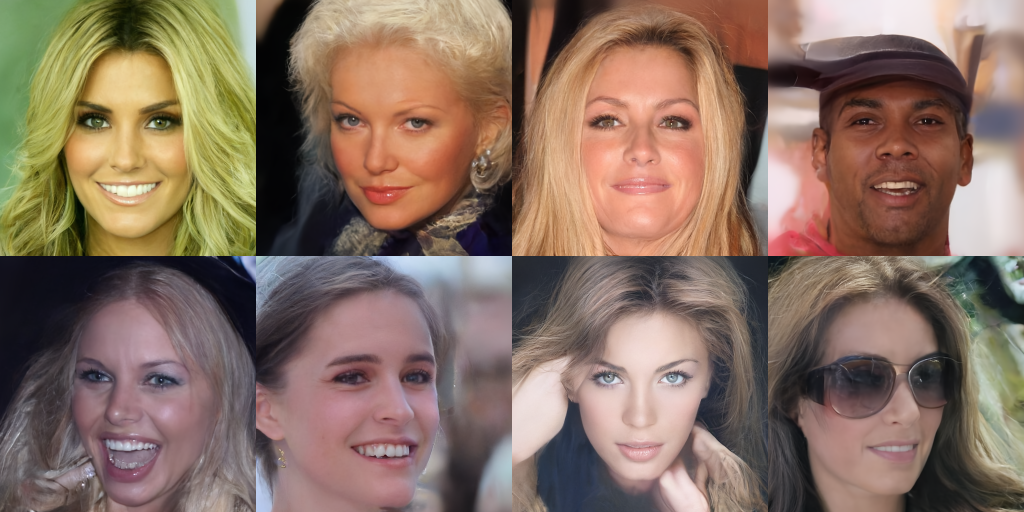}
        \captionsetup{font=scriptsize} 
        \caption{SA-2-DDIM $(T=50)$}
    \end{subfigure}
    \begin{subfigure}{0.48\textwidth}
        \centering
        \includegraphics[width=\linewidth]{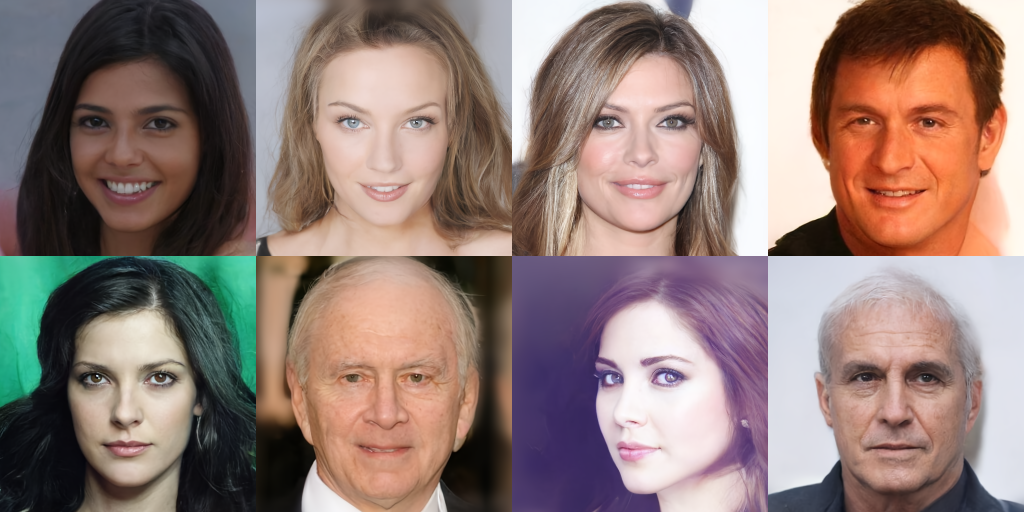}
        \captionsetup{font=scriptsize} 
        \caption{SA-2-DDPM $(T=200)$}
    \end{subfigure}
    \hspace{0.04\textwidth}%
    \begin{subfigure}{0.48\textwidth}
        \centering
        \includegraphics[width=\linewidth]{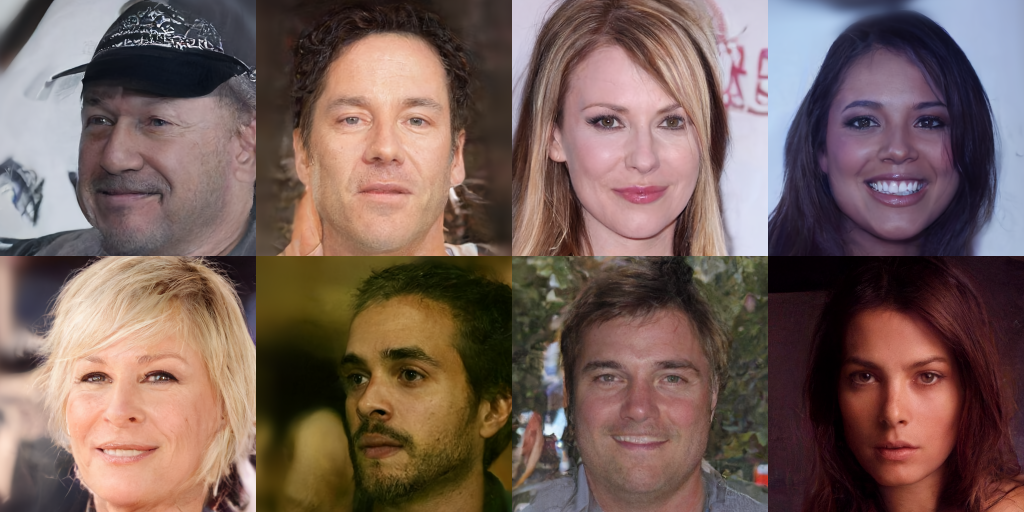}
        \captionsetup{font=scriptsize} 
        \caption{SA-2-DDIM $(T=200)$}
    \end{subfigure}
    \begin{subfigure}{0.48\textwidth}
        \centering
        \includegraphics[width=\linewidth]{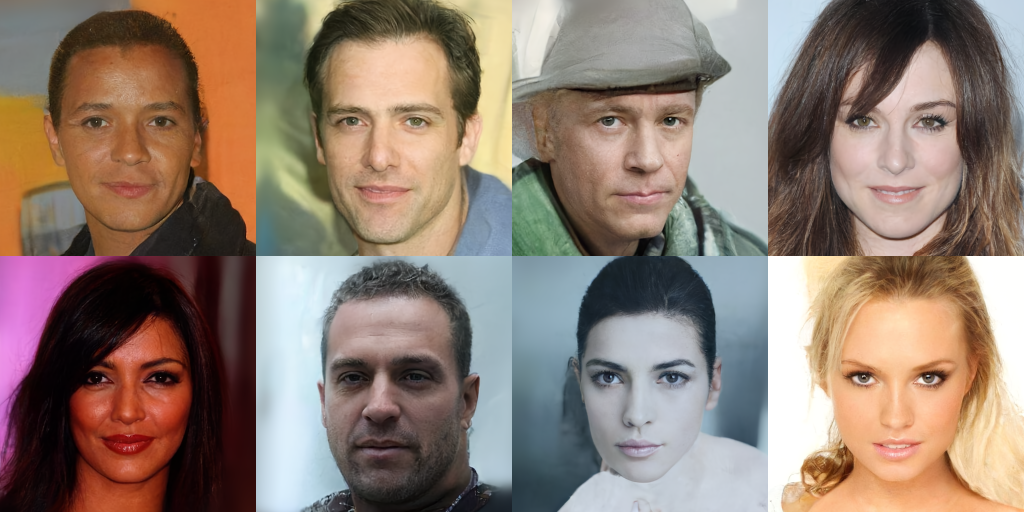}
        \captionsetup{font=scriptsize} 
        \caption{SA-2-DDPM $(T=1000)$}
    \end{subfigure}
    \hspace{0.04\textwidth}%
    \begin{subfigure}{0.48\textwidth}
        \centering
        \includegraphics[width=\linewidth]{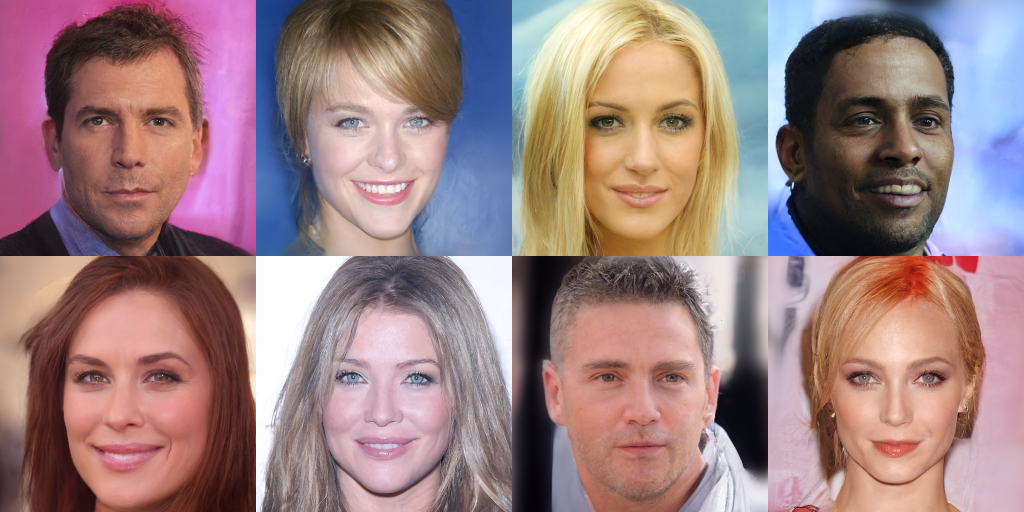}
        \captionsetup{font=scriptsize} 
        \caption{SA-2-DDIM $(T=1000)$}
    \end{subfigure}
    \caption{Generated samples on CelebA-HQ $256 \times 256$.}
    \label{fig:celebahq_sa_2}
\end{figure*}

Figures \ref{fig:cifar_t10}, \ref{fig:cifar_t50} and \ref{fig:cifar_t1000} report some samples of SA-$K$-DDPM and Vanilla-DDPM on trajectories of different number of timesteps $T$ on CIFAR10 dataset. \\
Figures \ref{fig:celeba_sa_2} and \ref{fig:celebahq_sa_2}  report some samples of SA-2-DDPM and SA-2-DDIM on trajectories of different number of timesteps $T$ on CelebA $64 \times 64$ and CelebA-HQ $256 \times 256$ dataset, respectively. Those figures should facilitate the comparison of image quality from different angles, such as training loss and timesteps.

\end{document}